\documentclass{article}

\usepackage{PRIMEarxiv}
\usepackage[utf8]{inputenc} 
\usepackage[T1]{fontenc}    
\usepackage{hyperref}       
\usepackage{url}            
\usepackage{booktabs}       
\usepackage{amsfonts}       
\usepackage{nicefrac}       
\usepackage{microtype}      
\usepackage{lipsum}
\usepackage{fancyhdr}       
\usepackage{graphicx}       
\graphicspath{{media/}}     
\usepackage{caption} 
\usepackage{threeparttable}
\usepackage{amsmath}
\usepackage{subcaption}
\usepackage{multirow}
\usepackage{makecell}
\usepackage[T1]{fontenc}
\usepackage[utf8]{inputenc}
\usepackage{authblk}

\DeclareMathOperator*{\argmin}{arg\,min}
\title{Formal Analysis of Art: Proxy Learning of Visual Concepts from Style Through Language Models\thanks{This paper is an extended version of a paper that will be published on the 36th AAAI Conference on Artificial Intelligence, to be held in Vancouver, BC, Canada, February 22 - March 1, 2022}}

\author[1]{\textbf{Diana S. Kim}}
\author[1]{\textbf{Ahmed Elgammal}}
\author[2]{\textbf{Marian Mazzone}}
\affil[1]{Department of Computer Science, Rutgers University, USA}
\affil[2]{Department of Art and Architectural History, College of Charleston, USA}
\affil[1]{dsk101@rutgers.edu, elgammal@cs.rutgers.edu}
\affil[2]{mazzonem@cofc.edu}

\begin{document}
\maketitle
\begin{abstract}
\par We present a machine learning  system that can quantify fine art paintings with a set of visual elements and principles of art. This formal analysis is fundamental for understanding art, but developing such a system is  challenging. Paintings have high visual complexities, but it is also difficult to collect enough training data with direct labels. To resolve these practical limitations, we introduce a novel mechanism, called \textit{proxy learning}, which learns visual concepts in paintings though their general relation to styles. This framework does not require any visual annotation, but only uses style labels and a general relationship between visual concepts and style. In this paper, we propose a novel proxy model and reformulate four pre-existing  methods in the context of proxy learning.  Through quantitative and qualitative comparison, we evaluate these methods and compare their effectiveness in quantifying the artistic visual concepts, where the general relationship is estimated by language models; GloVe \cite{pennington2014glove} or BERT \cite{devlin2018bert,vaswani2017attention}. The language modeling is a practical and scalable solution requiring no labeling, but it is inevitably imperfect. We demonstrate how the new proxy model is robust to the imperfection, while the other  models are sensitively affected by it.
\end{abstract}
\section{Introduction}
\par Artists and art historians usually use elements of art, such as line, texture, color, and shape \cite{fichner2011foundations}, and principles of art, such as balance, variety, symmetry, and proportion \cite{ocvirk2002art} to visually describe artworks. These elements and principles provide structured grounds for effectively communicating about art, especially the first principle of art, which is ``visual form'' \cite{ van1887principles}. 
\par However, in the area of AI, understanding art has mainly focused on a limited version of the first principle, through developing systems such as  predicting  styles \cite{elgammal2018shape,diana2018artprinciple}, finding  non-semantic features for style \cite{mao2017deepart}, or designing digital filters to extract some visual properties like brush strokes, color, textures, and so on \cite{berezhnoy2005computerized,johnson2008image}. While they are useful, the  concepts do not reveal much about the visual properties of paintings in depth. Kim et al. \cite{diana2018artprinciple} suggested a list of 58 concepts that break down the elements and principles of art. We focus on developing an AI system that can quantify such concepts. These concepts are referred to as ``visual elements'' in this paper and presented in Table \ref{tab:tab_0}.
\begin{table}[h!]
\centering
\resizebox{0.67\columnwidth}{!}{
\begin{tabular}{cc} 
\toprule
\cmidrule(r){1-2}
Elements and Principles of Art &  Concepts \\
\midrule
Subject&representational, non-representational\\\hline
Line&\makecell{blurred,
broken,
controlled\\
curved,
diagonal,
horizontal\\
vertical,
meandering\\
thick,
thin,
active,
energetic,
straight} \\\hline
Texture&\makecell{bumpy,
flat,
smooth\\
gestural,
rough
}\\\hline
Color&\makecell{calm,
cool,
chromatic\\
monochromatic,
muted,
warm,
transparent}\\\hline
Shape&\makecell{ambiguous,
geometric,
amorphous\\
biomorphic,
closed,
open,
distorted,
heavy\\
linear,
organic,
abstract,
decorative,
kinetic,
light}\\\hline
\makecell{Light and Space}&\makecell{bright,
dark, atmospheric \\
planar,
perspective}\\\hline
\makecell{General
Principles\\of Art} &\makecell{overlapping,
balance,
contrast\\
harmony,
pattern,
repetition,
rhythm\\
unity,
variety,
symmetry,
proportion,
parallel}\\         
\bottomrule
\end{tabular}}
\caption{A list of 58 concepts describing  elements and principles of art. We propose an AI system that can quantify such concepts. These concepts are referred to as \textbf{``Visual Elements"} in this paper.}

\label{tab:tab_0}
\end{table}

\par The main challenge in learning the visual elements and principles of art is that it is not easy to deploy any supervised methodology.  In general, it is difficult to collect enough annotation with multiple attributes. When it comes to art, the lack of visual element annotation becomes a more significant issue. Art  is typically annotated with artist information (name, dates, bio), style, genre attributes only, while annotating elements of art requires high specialties to identify the visual proprieties of artworks. Perhaps the sparsity of art data might be a reason  why art has been  analyzed computationally in the limited way.

\par To resolve the sparsity issue, this paper proposes to learn the visual elements of art through their general relations to styles (period style). While it is difficult to obtain the labels for the visual concepts, there are plenty of available paintings labeled by  styles and language resources  relating styles to visual concepts, such as online encyclopedia and museum websites. In general, knowing the dominant visual features of a painting enables us to identify its plausible styles. So we have the following questions; (1) what if we can take multiple styles as proxy components to encode visual information of paintings?  (2) Can a deep Convolutional Neural Network (deep-CNN) help to retrace visual semantics from the proxy representation of multiple styles? 

\par In these previous studies \cite{elgammal2018shape,diana2018artprinciple},  existence of the conceptual ties between visual elements and styles is demonstrated by using a hierarchical structure in the deep-CNN. They showed the machine can learn underlying semantic factors of styles from its hidden layers. Inspired by the studies, we hypothetically  set   a linear  relation between visual elements and style. Next, we constrain a deep-CNN  by the linear relation to make the machine learn visual concepts from its last hidden layer, while it is trained as a style classifier only. \par To explain the  methodology, a new concept--proxy learning--is defined first. It refers to all possible learning methods aiming to quantify paintings with a set of finite visual elements, which has no available label, by correlating it to another concept  that has abundant labeled data. In this paper, we reformulate four pre-existing methods in the context of proxy learning and introduce a novel approach that utilizes a deep-CNN to learn visual concepts from styles labels and language models. We propose to name it \textit{deep-proxy}. The output of deep-proxy quantifies the relatedness of an input painting to each of visual elements. In Table \ref{tab:tab0_0_1}, the most relevant or irrelevant visual elements are listed for the example paintings. The results are computed based on a deep-proxy model which is trained by only using the style labels and the language model, BERT
\par In the experiment, deep-proxy and four methods in attribute learning—sparse coding \cite{efron2004least}, logistic regression (LGT) \cite{danaci2016low}, Principal Component Analysis method (PCA) \cite{diana2018artprinciple}, and an Embarrassingly Simple approach to Zero-Shot Learning (ESZSL)  \cite{romera2015embarrassingly}—are quantitatively compared with each other. We analyze their effectiveness depending  two practical  solutions to estimate  a general relationship: (1) language models—GloVe  \cite{pennington2014glove} and BERT \cite{devlin2018bert,vaswani2017attention}—and (2) sample means of a few ground truth values.  
The language modeling is a practical and scalable solution requiring no labeling, but it is inevitably imperfect. We demonstrate how deep-proxy's cooperative structure  learning with styles creates strong resilience to the imperfection from the language models, while PCA and ESZSL are significantly affected by them. On the other hand, as a general relation is estimated by some ground truth samples, PCA performs best in various experiments. We summarize our contributions as follows.

\begin{enumerate}
    \item Formulating the proxy learning methodology and applying it to learn visual artistic concepts.
    \item A novel and end-to-end  framework to learn multiple visual elements from fine art paintings without any direct annotation. 
    \item A new word embedding trained by BERT \cite{devlin2018bert,vaswani2017attention}) and  a huge art corpus ($\sim 2,400,000$ sentences). This is a first BERT model for art,  trained by art-related texts. 
    \item A ground truth set of 58 visual semantics for 120 fine art paintings completed by seven art historians.
\end{enumerate}
\begin{table*}[t!]
\centering
\resizebox{0.95\textwidth}{!}{

  \begin{tabular}{m{1.0 in}|c||c|m{1.0 in}|c||c}
  \hline
\multirow{3}{*}{}&&&&&\\
Image&Relatedness&Words&Image&Relatedness& Words\\
&&&&&\\\hline
\multirow{8}{*}{\includegraphics[height=1.0in, width=1.0in]{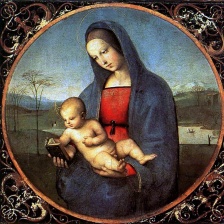}}&\multirow{4}{*}{\makecell{ R}}&&\multirow{8}{*}{\includegraphics[height=1.0in, width=1.0in]{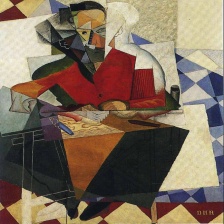}}&\multirow{4}{*}{\makecell{ R}}&\\
&&\makecell{\textbf{muted}, balance, representational}&&&\makecell{ \textbf{abstract}, blurred, transparent  }\\
&&\makecell{atmospheric, smooth}&&&\makecell{non-representational, thick}\\
&&&&&\\\cline{2-3}\cline{2-3}\cline{5-6}
&\multirow{4}{*}{\makecell{ IR}}&&&\multirow{4}{*}{\makecell{ IR}}&\\
&&\makecell{planar, rhythm, blurred }&&&\makecell{dark, horizontal, controlled }\\
&&\makecell{thick, \textbf{abstract}}&&&\makecell{balance, \textbf{representational}}\\
&&&&&\\\hline\hline

\multirow{8}{*}{\includegraphics[height=1.0in, width=1.0in]{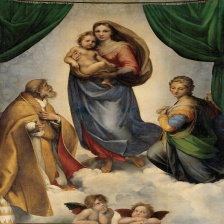}}&\multirow{4}{*}{\makecell{ R}}&&\multirow{8}{*}{\includegraphics[height=1.0in, width=1.0in]{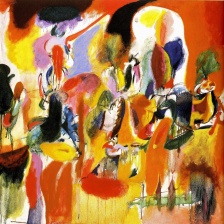}}&\multirow{4}{*}{\makecell{ R}}&\\
&&\makecell{ \textbf{muted}, balance, heavy }&&&\makecell{\textbf{abstract}, blurred thick }\\
&&\makecell{controlled, representational }&&&\makecell{non-representational, biomorphic }\\
&&&&&\\\cline{2-3}\cline{2-3}\cline{5-6}
&\multirow{4}{*}{\makecell{ IR}}&&&\multirow{4}{*}{\makecell{ IR}}&\\
&&\makecell{planar, rhythm, thick }&&&\makecell{balance, smooth, planar }\\
&&\makecell{abstract, \textbf{blurred}}&&&\makecell{dark, \textbf{representational}}\\
&&&&&\\\hline\hline

\multirow{8}{*}{\includegraphics[height=1.0in, width=1.0in]{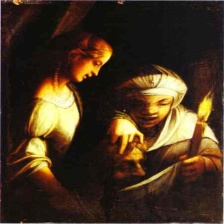}}&\multirow{4}{*}{\makecell{ R}}&&\multirow{8}{*}{\includegraphics[height=1.0in, width=1.0in]{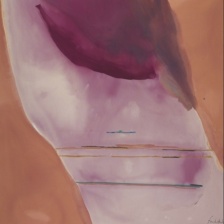}}&\multirow{4}{*}{\makecell{ R}}&\\
&&\makecell{\textbf{dark}, atmospheric, muted}&&&\makecell{\textbf{abstract}, blurred, thick  }\\
&&\makecell{horizontal, representational }&&&\makecell{biomorphic, non-representational}\\
&&&&&\\\cline{2-3}\cline{2-3}\cline{5-6}
&\multirow{4}{*}{\makecell{ IR}}&&&\multirow{4}{*}{\makecell{ IR}}&\\
&&\makecell{amorphous, rhythm, thick}&&&\makecell{rough, kinetic, balance }\\
&&\makecell{blurred, \textbf{planar}}&&&\makecell{smooth, \textbf{representational}}\\
&&&&&\\\hline\hline

\multirow{8}{*}{\includegraphics[height=1.0in, width=1.0in]{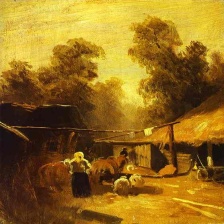}}&\multirow{4}{*}{\makecell{R}}&&\multirow{8}{*}{\includegraphics[height=1.0in, width=1.0in]{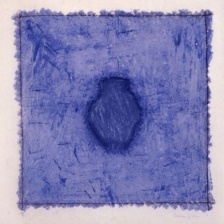}}&\multirow{4}{*}{\makecell{ R}}&\\
&&\makecell{\textbf{atmospheric}, dark, muted  }&&&\makecell{ \textbf{abstract}, thick, biomorphic }\\
&&\makecell{horizontal, warm}&&&\makecell{gestural, pattern}\\
&&&&&\\\cline{2-3}\cline{2-3}\cline{5-6}
&\multirow{4}{*}{\makecell{IR}}&&&\multirow{4}{*}{\makecell{ IR}}&\\
&&\makecell{planar, thick, kinetic}&&&\makecell{geometric, amorphous, monochromatic}\\
&&\makecell{rough, \textbf{amorphous}}&&&\makecell{ planar, \textbf{representational}}\\
&&&&&\\\hline\hline

\multirow{8}{*}{\includegraphics[height=1.0in, width=1.0in]{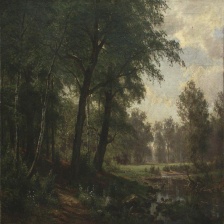}}&\multirow{4}{*}{\makecell{ R}}&&\multirow{8}{*}{\includegraphics[height=1.0in, width=1.0in]{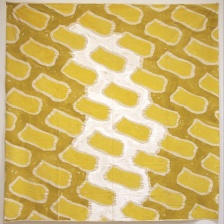}}&\multirow{4}{*}{\makecell{ R}}&\\
&&\makecell{\textbf{atmospheric}, dark, smooth}&&&\makecell{\textbf{abstract}, thick, blurred}\\
&&\makecell{warm, muted}&&&\makecell{biomorphic, rhythm }\\
&&&&&\\\cline{2-3}\cline{2-3}\cline{5-6}
&\multirow{4}{*}{\makecell{ IR}}&&&\multirow{4}{*}{\makecell{ IR}}&\\
&&\makecell{thick, kinetic, rough}&&&\makecell{balance, controlled, smooth}\\
&&\makecell{  planar, \textbf{amorphous}}&&&\makecell{ dark, \textbf{representational}}\\
&&&&&\\\hline\hline
\end{tabular}}

\caption{The Relevant (R) and Irrelevant (IR) Visual Elements by Deep-Proxy: Based on the output of deep-proxy, top and bottom five ranked visual elements are listed. In this result, deep-proxy is trained by  using the style labels and the general relationship estimated by the language model, BERT. The most relevant or irrelevant words are in bold. The title, author, year of made, and style of these paintings are shown in Supplementary Information (SI) A.}
\label{tab:tab0_0_1}
\end{table*}

\section{Related Work}
\subsection{Attribute Classification}For learning semantic attributes, mainstream literature has been based on simple binary classification and fully  \cite{farhadi2009describing,lampert2013attribute} or weakly supervision methods \cite{ferrari2007learning,shankar2015deep}. Support Vector Machine \cite{farhadi2009describing,lampert2013attribute,patterson2014sun} and logistic regression \cite{danaci2016low,farhadi2009describing} are used to recognize the presence or absence of targeted semantic attributes. 

\subsection{Descriptions by Visual Semantics}
This paper's method is not designed using a classification problem, but rather it generates real-valued vectors.  Each dimension of each vector is aligned with a certain visual concept, so the vectors naturally indicate which paintings are more or less relevant to the concept. As is the case with most similar formats, Parikh et al. \cite{parikh2011relative,ma2012unsupervised} propose to predict the relative strength of the presence of attributes through real-valued ranks. 
\par  For attribute learning, recently its practical merits  have been  rather emphasized, such as zero-shot learning \cite {xian2018zero} and semantic \cite{li2010objects} or  non-semantic attributes \cite{huang2016unsupervised} to boost object recognition.  However, in this paper, we  focus on attribute learning itself and pursue its descriptive and human understandable
advantages, in the same way that Chen et al. \cite{chen2012describing} focused on describing  clothes with some words understandable to humans.
\subsubsection{Incorporating Classes as Learning Attributes}Informative dependencies between semantic attributes and  objects (class) are useful; in fact,  they have co-appeared  in many papers. Lampert et al. \cite{lampert2013attribute} assign attributes to images on a per-class basis and train attribute classifiers in a supervised way. On the other hand, Yu et al.  \cite{yu2014modeling} model attributes based on their generalized properties—such as their proportions and relative strength—with a set of categories and make learning algorithms  satisfy them as necessary conditions. The methods do not require any instance-level attributes for training like this paper method, but learning visual elements satisfying the constraints of relative proportions among classes is not related to our goal or methodology. Some researchers \cite{mahajan2011joint,wang2013unified} propose joint learning frameworks to more actively incorporate class information into attribute learning. In particular, Akata et al. \cite{akata2013label} and Romera-Paredes et al. \cite{romera2015embarrassingly}
hierarchically treat  attributes as intermediate features, which serve to describe classes. The systems are designed to learn attributes by bi-directional influences from class to attributes (top-down) and from image features to attributes (bottom-up) like deep-proxy. However, their single and linear layering, from image features to their intermediate attributes, are different from the multiple and non-liner layering in deep-proxy.
\subsubsection{Learning Visual Concepts from Styles}
Elgammal et al. \cite{elgammal2018shape,diana2018artprinciple} 
 show that a deep-CNN can learn semantic factors of styles from its last hidden layers by using a hierarchical structure of deep-CNN. They interpret deep-CNN's last hidden layer with pre-defined visual concepts through multiple and separated  post-procedures, but deep-proxy simultaneously learns visual elements while machines are trained for style classification. In the experiment, the method proposed by Kim et al. \cite{diana2018artprinciple} is compared with deep-proxy as the name of PCA. 

\section{Methodology}
\subsection{Linear Relation}
\subsubsection{Two Conceptual Spaces}
\begin{figure}[htbp]

\centering
\includegraphics[width=0.4\linewidth]{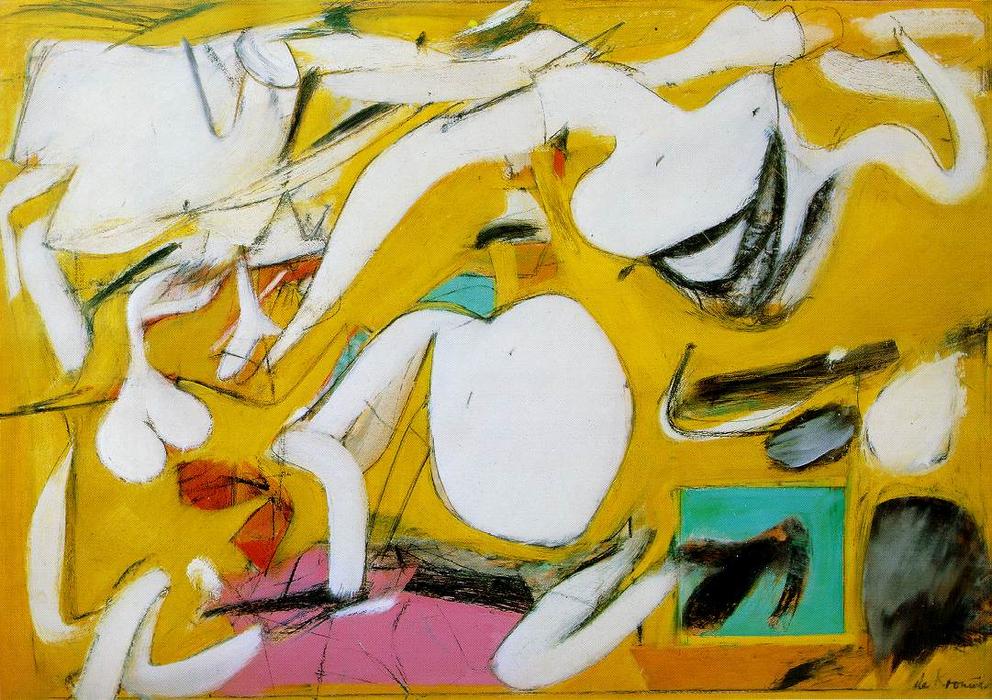}
\includegraphics[width=1.05\linewidth]{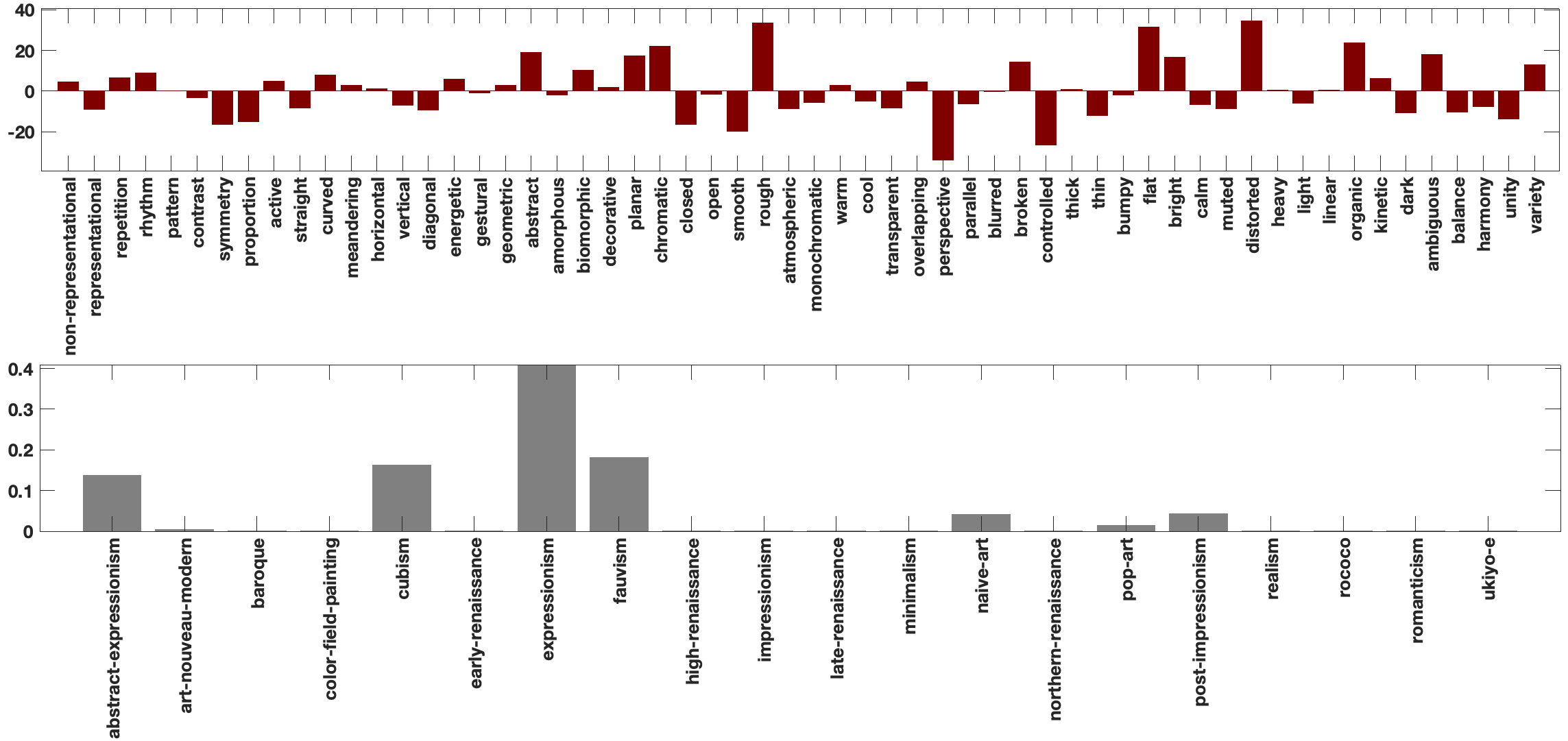}
\caption{Fire Island by Willem de Kooning (1946): Along with the original style of Abstract Expressionism, Expressionism, Cubism, and Fauvism have all left visual traces within the painting. These are earlier styles that De Kooning, the artist, knew and learned from before developing his own mature style. And the visual traces of the styles are as follows: the breaking apart of whole forms or objects (Cubism), a loose, painterly touch (Expressionism) and vivid, striking color choices and juxtapositions such as pink, teal and yellow (Fauvism).}
\label{fig:fig_0_1}
\end{figure}

\begin{figure}[htbp]

\centering
\includegraphics[width=0.375\linewidth]{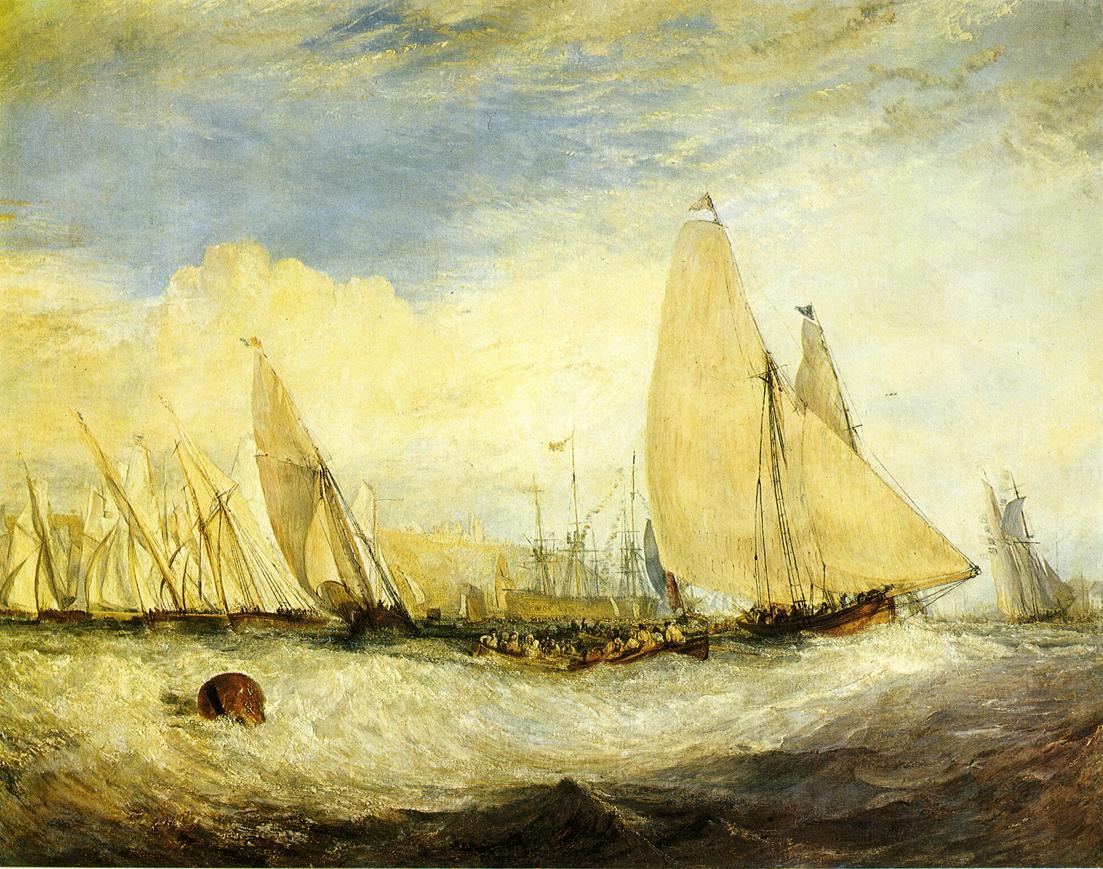}
\includegraphics[width=1.0\linewidth]{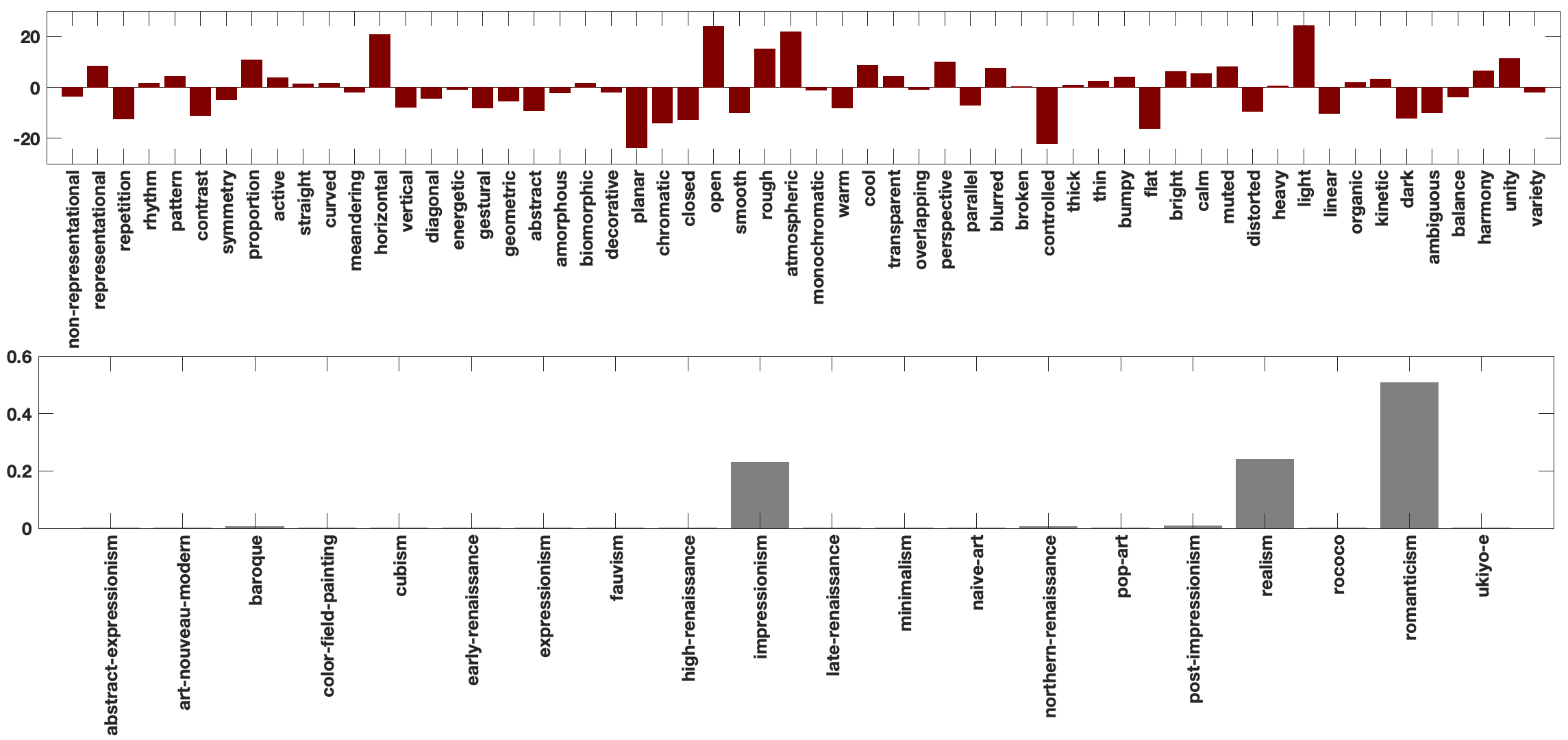}
\caption{East Cowes Castle, the seat of J.Nash, Esq. the Regatta beating to windward
by J.M.W Turner (1828): The
painting is originally belonged to Romanticism but Realism and Impressionism are also used for its style encoding.}

\label{fig:fig_0_2}
\end{figure}

Styles are seldom completely uniform and cohesive, and often carry forward within them former styles and other influences that are still operating within the work. As explained in \textit{The Concept of Style} \cite{lang1987concept},  a style can be both a possibility and an interpretation. It is not a definite quality that inherently belongs to objects, although each of the training samples are artificially labeled with a unique style. Due to the complex variations of the visual properties of art pieces in sequential arrangements of times, styles can be overlapped, blended, and merged.  Based on the idea, this research begins with representing  paintings with the two entities: a set  of $m$ visual elements and a  set  of $n$ styles. Two conceptual vector spaces $\mathbf{S}$ and $\mathbf{A}$ for style and visual elements are introduced, whose  each dimension  is aligned with their semantic. Two vector functions, $f_{A}(\cdot)$ and  $f_{S}(\cdot)$, are defined to transform  input image $x$ into the conceptual spaces in equation (\ref{eqn:eqn_1}) below. 
\begin{align}
\begin{split}
f_{A}(x): x \rightarrow \vec{a}(x)&=[a_{1}(x),a_{2}(x),...,a_{m}(x)]^{t}\in \mathbf{A}\\
f_{S}(x): x \rightarrow 
\vec{s}(x)&=[s_{1}(x),s_{2}(x),...,s_{n}(x)]^{t}\in \mathbf{S}
\label{eqn:eqn_1}
\end{split}
\end{align}
Figure \ref{fig:fig_0_1} and  \ref{fig:fig_0_2} show  example paintings that are encoded by visual elements and styles. They are generated by deep-proxy (using a general relationship estimated by sample means of a few ground truth value).

\subsubsection{Category-Attribute Matrix}
\par Inspired by a prototype theory \cite{murphy2004big} in cognitive science, we posit that a set of pre-defined visual elements of art is sufficient for characterizing  styles. According to this theory, once a set of attributes are arranged to construct a vector space, a vector point can summarize each of categories. Mathematically modeling the principles,  $x_{s_{i}^{*}}$ is set to be the typical (best) example  for the style $s_{i}$, where $i \in \{1,2,...n\}$ and $n$ is the number of styles. This is represented by 
$f_{A}(x_{s_{i}}^{*})=x \rightarrow \vec{a}(x_{s_{i}}^{*})$.
By accumulating the $\vec{a}(x_{s_{i}}^{*}) \in \mathbb{R}^{m} $ as columns for all different $n$ styles,  a matrix $G \in \mathbb{R}^{m \times n}$ is formed. Matrix $G$ becomes a category-attribute matrix.
\subsubsection{Linear System}
Matrix  $G$ ideally defines  $n$ typical points  for $n$ styles in the attribute space of $\mathbf{A}$. However, as aforementioned, images that belong to a specific style show intra-class variations. In this sense, for a painting $x$ that belongs to style $s_{i}$, the 
$f_{A}(x_{s_{i}}^{*})$
 is likely to be the closest  to the $f_{A}(x)$, and  its similarities  to other styles' typical points can  be calculated by the inner products between $f_{A}(x)$ and $f_{A}(x_{s_{i}}^{*})$ for all $n$ styles, $i \in \{1,2,...n\}$. All computations are expressed by $f_{A}(x)^{t}\cdot G$ and its output  $f_{S}(x)^{t}$. This results in the linear equation (\ref{eqn:eqn_2}) below.
\begin{equation}
f_{A}(x)^{t} \cdot G = f_{S}(x)^{t}
\label{eqn:eqn_2}
\end{equation}
\begin{figure*}[t!]
 \centering
  \includegraphics[width=1.05\linewidth]{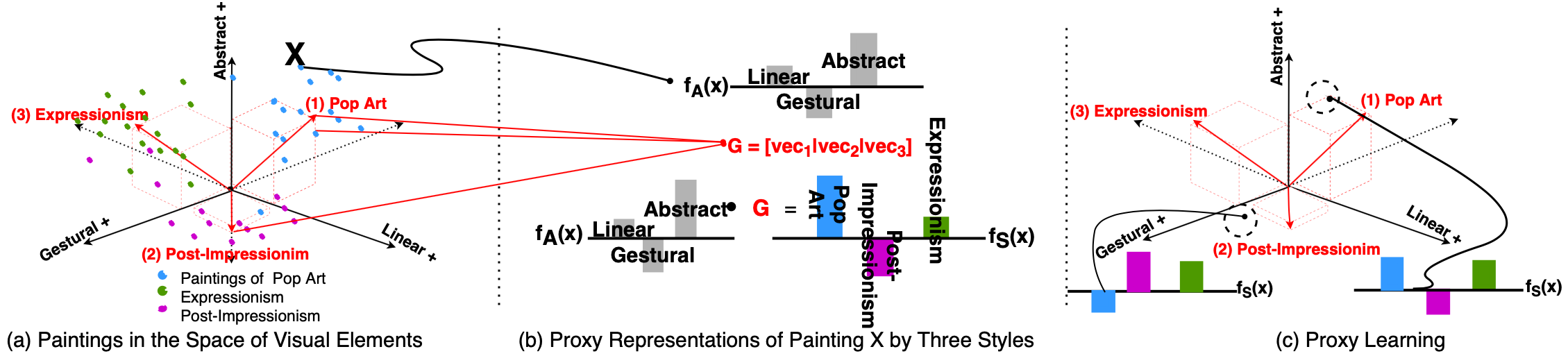}
\caption{Summary of Proxy Learning:  (a) The paintings of three styles (Pop art, Post-impressionism, and Expressionism) are scattered in the  space of three visual elements (abstract, gestural, and linear). The red vectors represent typical vectors of the three styles. (b) A painting $X$, originally positioned in the visual space, can be transformed to the three-style (proxy) representation by computing inner products with each of the typical vectors.
(c) Proxy learning aims to estimate or learn its original visual coordinates from a proxy representation and a set of typical vectors.} 
\label{fig:fig_1}
\end{figure*}
\subsection{Definition of Proxy Learning}
In equation (\ref{eqn:eqn_2}), knowing $f_{S}(\cdot)$ becomes linearly tied with knowing  $f_{A}(\cdot)$, so we have the following questions: (1) given $G$ and $f_{S}(\cdot)$, how can we learn the function  $f_{A} (\cdot)$? (2) before doing that, how can we properly encode  $G$ and  $f_{S}(\cdot)$ first? This paper aims to answer these questions. We first re-define them by a new concept of learning, named \textit{proxy learning}. Figure. \ref{fig:fig_1} is an illustrative example to describe it.\\

\par \textit{Proxy learning: a computational framework that learns the function  $f_{A} (\cdot)$ from $f_{S}(\cdot)$ through a linear relationship $G$. $G$ is estimated by language models or human survey.}
\subsection{Language Modeling}
The $G$ matrix is estimated by using distributed word embeddings in NLP. Two  embeddings were considered: GloVe  \cite{pennington2014glove} and BERT \cite{devlin2018bert,vaswani2017attention}. However, their original dictionaries do not provide all the necessary art terms. Especially for BERT, it holds a relatively smaller dictionary than  GloVe. In the original BERT,  vocabulary words are represented by several word-pieces \cite{wu2016google}, so it is unnecessary to  hold a large set of words.  However, the token-level vocabulary words could lose their original meanings, so a new BERT model had to be trained from scratch on a suitable art corpus in order to compensate for the deficient dictionaries. 
\subsubsection{A Large Corpus of Art}
To prepare a large corpus of art, we first gathered all the descendent categories  (about 6500) linked with the parent categories of \texttt{``Painting''} and \texttt{``Art Movement''} in Wikipedia and scrawled all the texts under the categories by using a library available in public. Some art terms  and their definitions presented in  TATE museum \footnote{http://tate.org.uk/art/art-terms} were also added, so finally, with $\sim 2,400,000$ sentences, a  set word embedding set—BERT—is newly trained for art. 

\subsubsection{Training BERT}
For a new BERT model for art, the BERT-BASE model (12-layer, 768-hidden, 12-heads, and not using cased letters) was selected and trained from scratch with the collected art corpus. For training, the original vocabulary set is updated by adding some words which are missed in the original framework.  We averaged all 12 (layers) embeddings to compute each of final word embeddings. All details about BERT training is presented in SI B.
\subsubsection{Estimation of Category-Attribute Matrix  $G$}
To estimate a matrix $G$,  vector representations were collected  and  the following over-determined system of equations was set. 
Let the $W_{A} \in \mathbb{R}^{d \times m}$ denote a matrix  of which each column implies a $d$-dimensional word embedding to encode one of $m$ visual elements, and the  $w_{s_{i}} \in \mathbb{R}^{d}$ be a word embedding that represents  style $s_{i}$ among $n$ styles. 
\begin{equation}
W_{A} \cdot \vec{a}(x_{s_{i}}^{*}) = w_{s_{i}}
\label{eqn:eqn_3}
\end{equation}
By solving the equation (\ref{eqn:eqn_3}) for  $i \in \{1,2,...,n\}$,  the vector $\vec{a}(x_{s_{i}}^{*}) \in \mathbb{R}^{m}$  was estimated, which becomes each  column vector of  ${G}$. It quantifies how the visual elements are positively or negatively related to a certain style in a distributed vector space. In general, word embedding geometrically captures semantic or syntactical relations between words, so this paper postulates that the general relationship among the concepts can be reflected by the linear formulation (\ref{eqn:eqn_3}).
\subsection{Deep-Proxy}
\begin{figure}[h]
  \centering
  \includegraphics[width=0.85\linewidth]{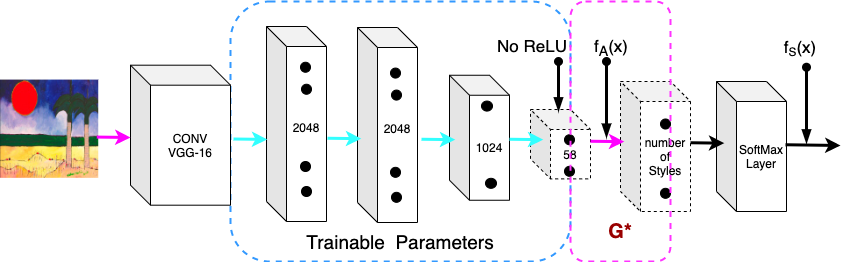}
\caption{Deep-Proxy Architecture}
\label{fig:fig_2}
\end{figure}
We propose a novel method to jointly learn the two multi-modal functions, $f_{S}(x)$ and $f_{A}(x)$, through a pre-estimated general matrix ($G$).  Its principal mechanism is a category-attribute matrix ($G$) is hard-coded into the last fully connected (FC) parameters of a deep-CNN, so it is enforced to learn visual elements ($f_{A}(x)$) from its last hidden layers, while it is outwardly trained to learn multiple styles ($f_{S}(x)$). We propose to name this framework \textit{deep-proxy}.  In this paper, the original VGG-$16$ \cite{Simonyan15} is adapted for its popularity and modified as a style classifier, as shown in Figure \ref{fig:fig_2}. 
\subsubsection{Implementation of Deep-Proxy} All convolution layers are transferred from the ImageNet as is and frozen, but the original FC layers, $(4096 - 4096 - 1000)$, are expanded to the five layers $(2048 -  2048  - 1024 - 58 - G^{*} (58 \times n) - n \text{ number of styles})$. These FC parameters (cyan colored and dashed box) are updated during training.
We also tried to fine-tune convolution parts, but they showed slightly degraded performance compared to  the FC-only training. Therefore, all the results  presented in the later sections are FC-only trained for $200,000$ steps at $32$ batch size by the momentum optimizer (momentum $= 0.9$). The learning rate is initially set as $1.0e-3$ and degraded at the factor of $0.94$ every $2$ epochs.
The final soft-max output is designed  to encode the $f_{S}(x)$, and the last hidden layer ($58$-D) is set to encode the $f_{A}(x)$. The two layers are  interconnected by the FC block $G^{*}$ (magenta colored and dashed box) to impose a linear constraint between the two modalities. For the $f_{A}(x)$, the hidden layer's Rectified Linear Units (ReLU) is removed, so it can have both positive and negative values. 
\subsubsection{Objective Function of Deep-Proxy}
\par Let  $I (q,k)$ be an indicator function  stating whether or not the $k$-th sample belongs to  style class $q$. Let  $s_{q}(x|w)$ be the $q$-th style component of the soft-max simulating $f_{S}(x)$. Let  $f_{A}(x|w)$ be the last hidden activation vector, where  $x$ is an input image and $\omega$ is the network's parameters. Then, an objective for  multiple style classification is set as in equation (\ref{eqn:eqn_4}) below. The  $\lambda$ is added to regularize the magnitudes of the last hidden layer. 
\par \begin{equation}
\min_{\omega} \sum_{k}^{K} \sum_{q}^{Q} -I(q,k) \cdot \log_{e}(s_{q}(x|\omega)) + \lambda \cdot \lVert{f_{A}(x|\omega)}\rVert_{1} 
\label{eqn:eqn_4}
\end{equation}
In the next subsections, three versions of deep-proxy are defined depending on how  $G^{*}$ matrix is formulated. 
\subsubsection{Plain Method $(G^{*} = G)$}
A  $G$ matrix is estimated and plugged into the network as it is. Two practical solutions are considered to estimate  $G$, language models and sample means of a few ground truth values. In training for Plain, the $G^{*}$ is fixed as the $G$ matrix, while the other FC layers are updated. This modeling is exactly aligned with  equation (\ref{eqn:eqn_2}).
\subsubsection{SVD Method}
A structural disadvantage of  Plain method is noted and resolved by using Singular Vector Decomposition (SVD). 
\par It is natural that the columns of a $G$ matrix are correlated because a considerable number of visual properties are shared among  typical styles. Thus, if the machine learns visual space properly, the multiplication of $f_{A}(x)^{t}$ with $G$ necessarily produces $f_{S}(x)^{t}$, which is highly valued on multiple styles. On the other hand, the deep-proxy is trained by one-hot vectors  promoting orthogonality among styles and a sharp high value on a specific style component. Hence, learning with one-hot vectors can cause interference on learning visual semantics if we simply adopt the plain method above. For example, suppose there is a typical Expressionism painting $x_{*}$. then, it is likely to be highly valued both on the Expressionism and Abstract-Expressionism under the equation (\ref{eqn:eqn_2}) because the two styles are  correlated visually. But if  one hot-vector encourages the machine to value $f_{S}(x_{*})$ highly on the Expressionism axis only, then the machine might not be able to learn visual concepts well, such as gestural brush-strokes or mark-making, and the impression of spontaneity, because those concepts are supposed to be high on Expressionism, too. To fix this discordant structure, the $G$ and $f_{A}(x)$ are transformed to the space where typical style representations are orthogonal to one another. It reformulates equation (\ref{eqn:eqn_2}) by the equation (\ref{eqn:eqn_5}), where $T$ is a transform matrix to the space.
\begin{equation}
[f_{A}(x)^{t}\cdot T^{t}] \cdot [T  \cdot G] =  f_{S}(x)^{t} 
\label{eqn:eqn_5}
\end{equation}
\par To compute the transform matrix $T$,  $G$ is  decomposed by SVD.  As the number of attributes ($m$) is greater than  the number of classes ($n$), and its rank is $n$,  the $G$ is decomposed by $U \cdot \Sigma \cdot V^{t}$, where  the $U (m \times n)$ and $V (n \times n)$ are the matrices whose columns are orthogonal and the $\Sigma$ $(n \times n)$ is a diagonal matrix. From the decomposition,   $V^{t} =  \Sigma^{-1} \cdot U^{t} \cdot G$, so we can use $\Sigma^{-1} \cdot U^{t}$ as the transform matrix $T$. The $T = \Sigma^{-1} \cdot U^{t}$ transforms each column of $G$ to each orthogonal column of $V^{t}$. In this deep-proxy SVD method, the $G^{*}$ is reformulated by these SVD components as presented in equation (\ref{eqn:eqn_6}) below.
\begin{equation}
G^{*} = T ^{t} \cdot T \cdot G = U \cdot \Sigma^{-2} \cdot U^{t} \cdot G
\label{eqn:eqn_6}
\end{equation}
\subsubsection{Offset Method}
A positive offset vector $\vec{o}\in R^{+m}$ is introduced to learn a threshold to determine a visual concept as relevant or not. Each component of $\vec{o}$ implies an individual threshold for each of the visual elements, so when it is subtracted from a column of a $G$ matrix, we can take zero as an absolute threshold to interpret whether or not visual concepts are relevant to a style class. Since  matrix $G$ is often encoded by the values between zero and one, especially when it is created by human survey (ground truth values), we need a proper offset to shift the $G$ matrix. Hence, the vector $\vec{o}\in R^{+m}$ is set as learnable parameters in the third version of deep-proxy. It sets the $G^{*}$  as $U \cdot \Sigma^{-2} \cdot U^{t} \cdot (G - \boldsymbol \mu )$, where 
$\boldsymbol\mu=[\vec{o}|\vec{o}|...|\vec{o}]$ is the tiling matrix of the vector $\vec{o}$. In  Offset method, the SVD components $U$ and $\Sigma$ are newly calculated for the new $(G - \boldsymbol \mu )$ at every batch in training.

\section{Experiments}
Four pre-existing  methods, sparse coding \cite{efron2004least}, logistic regression (LGT)  \cite{danaci2016low}, Principal Component Analysis (PCA) method \cite{diana2018artprinciple}, and an Embarrassingly Simple approach to Zero-Shot Learning (ESZSL)  \cite{romera2015embarrassingly} are reformulated in the context of proxy learning and  quantitatively compared with each other. In this section, we  demonstrate how LGT and deep-proxy based on deep-learning are more robust than  others when a general relationship ($G$ matrix) is estimated by language models; GloVe \cite{pennington2014glove} or BERT \cite{devlin2018bert,vaswani2017attention}. We also show  LGT is degraded sensitively, when $G$ matrix is  sparse. All detailed implementations of the four pre-existing methods are explained in SI C. 
\subsection{Four Proxy Methods}
\subsubsection{Logistic Regression (LGT) \cite{danaci2016low}} Each column of $G$ was used to assign attributes to images on a per class basis.  When $G$ matrix  is not a probabilistic representation, without shifting zero points, the positives were put into the range of $0.5$ to $1.0$ and the negatives were put into the range of $0.0$ to $0.5$. 
\subsubsection{PCA \cite{diana2018artprinciple}} The last hidden feature of a deep-CNN style classifier is encoded by  styles and then multiplied with the transpose of $G$ matrix to finally compute each degree of the visual elements.
\subsubsection{ESZSL \cite{romera2015embarrassingly}}
This can be regarded as a special case of the deep-proxy Plain by setting a single  FC  layer between  visual features and $f_{A}(x)$, replacing the softmax loss with  Frobenius norm $\lVert{\cdot}\rVert_{Fro}^{2}$, and encoding styles with $\{-1,+1\}$. To  compute  the single  layer,  a global  optimum  is found through a closed-form formula proposed by Romera-Paredes et al. \cite{romera2015embarrassingly}.
\subsubsection{Sparse Coding \cite{efron2004least}} It estimates $f_{A}(\cdot)$ directly from the style encodings $f_{S}(\cdot)$ and $G$ matrix by solving a sparse coding equation without seeing input images. Its  better performance versus random cases proves our hypothetical modeling assuming informative ties between style and visual elements. 
\subsection{WikiArt Data Set and Visual Elements} 
This paper used the $76921$ paintings in  WikiArt's data set \cite{wikiart} and merged their original $27$ styles into $20$  styles \footnote{ Abstract-Expressionism, Art-Nouveau-Modern, Baroque, Color-Field-Painting,  Cubism, Early-Renaissance, Expressionism, Fauvism, High-Renaissance, Impressionism, Mannerism, Minimalism, Na\"{i}ve-Art-Primitivism, Northern-Renaissance, Pop-Art, Post-Impressionism, Realism, Rococo, Romanticism, and Ukiyo-e}, the same as those presented by Elgammal et al. \cite{elgammal2018shape}. $120$ paintings were separated for evaluation, and the remaining samples were randomly split into $85\%$ for training and $15\%$ for validation.  This paper adopts the pre-selected  visual concepts proposed by Kim et al. \cite{diana2018artprinciple}.  In the  paper, $59$ visual words are suggested, but we used $58$ words, excluding the  ``medium'' because it is not descriptive.

\subsection{Evaluation Methods}
\subsubsection{Human Survey}
A binary ground truth set was completed by seven art historians. The  subjects were asked to choose between one of the following three choices: (1) yes, the shown attribute and painting are  related. (2) they are somewhat relevant. (3) no, they are not related. Six paintings were randomly selected from each of 20 styles, and art historians made three sets of ground truths of 58 visual elements for the 120 paintings first.  From the three sets, a set was determined based on the majority vote. For example, if a question is marked by three different answers, the (2) `somewhat relevant' is determined as the final answer. The results show  1652 (24\%) as relevant, 782 (11\%) as somewhat, and 4526 (65\%) as irrelevant.  In order to balance positive  and negative  values,  this paper considered the somewhat answers as relevant and created a binary ground truth set.  The 120 paintings  will be called ``eval'' throughout this paper.

\subsubsection{AUC Metric} 
The \textbf{A}rea \textbf{U}nder the receiver operating characteristic \textbf{C}urve (AUC) was used for evaluation. When we say AUC$@K$, it means an averaged AUC score, where the $K$ denotes the number of attributes to be averaged. A random case is simulated and drawn in every  plot as a comparable baseline. Images are sorted randomly without the consideration of the machine’s out values and then AUCs are computed. We explained why AUC is selected for art instead of other metrics (mAP or PR) in SI D. 
\subsubsection{Plots} To draw a plot, we measured $58$ AUC scores for all $58$ visual elements.  The scores were sorted in descending order,  every three scores were grouped, and  $19$ $(\lfloor58/3\rfloor)$ points of AUC$@3$ were computed. Since many  of the  visual concepts were not learnable (AUC $\leq$ $0.5$), a single averaged AUC$@58$ value did not differentiate  performance clearly. Hence, the descending scores were used, but averaged at every three points for simplicity. Regularization parameters were written in the legend boxes of plots if necessary. 
\subsubsection{SUN and CUB}
SUN \cite{patterson2014sun} and CUB \cite{WahCUB_200_2011} are used to understand the models in  general situations. All experiments are based on the standard splits, proposed by Xian et al. \cite{xian2018zero}. For evaluation, mean Average Precision (AP) is used because the ground truth of the data sets is imbalanced by very large negative samples (the mean  of all the samples is $0.065$ for SUN and $0.1$ for CUB at the binary threshold of $0.5$). For $G$ matrix, their ground truth samples are averaged.
\subsection{Estimation of Category-Attribute Matrix}
Two ways to estimate $G$ matrix are considered. First,  from the two sets of  word embeddings—GloVe and BERT—two $G$ matrices are computed by equation (\ref{eqn:eqn_3}).  This paper will refer to   the BERT matrix as  $G_{B}$  and  to  the GloVe matrix as $G_{G}$. The $G_{G}$ is used only for $12$-style experiments in a section later because the vocabulary of GloVe  does not contain the all terms for the $20$-style. As necessary, they will be written with the number of styles involved in experiments like $G_{B_{20}}$ or $G_{B_{12}}$. Second, $58$-D ground truths of the three paintings, randomly selected among the six paintings of each style, were averaged and accumulated into columns, and the ground truth matrix $G_{GT}$ was also established.  To do so, we first mapped the three answers of the survey with integers: ``relevant'' = $+1$; ``somewhat'' = $0$; and ``irrelevant'' = $-1$. The $60$ paintings of ``eval'' used to compute  $G_{GT}$ will be called ``eval-$G$'' and the others will be called ``eval-$NG$''.

\subsection{Quantitative Evaluations}
\begin{figure*}[t!]
 \centering
  \includegraphics[width=1.0\linewidth]{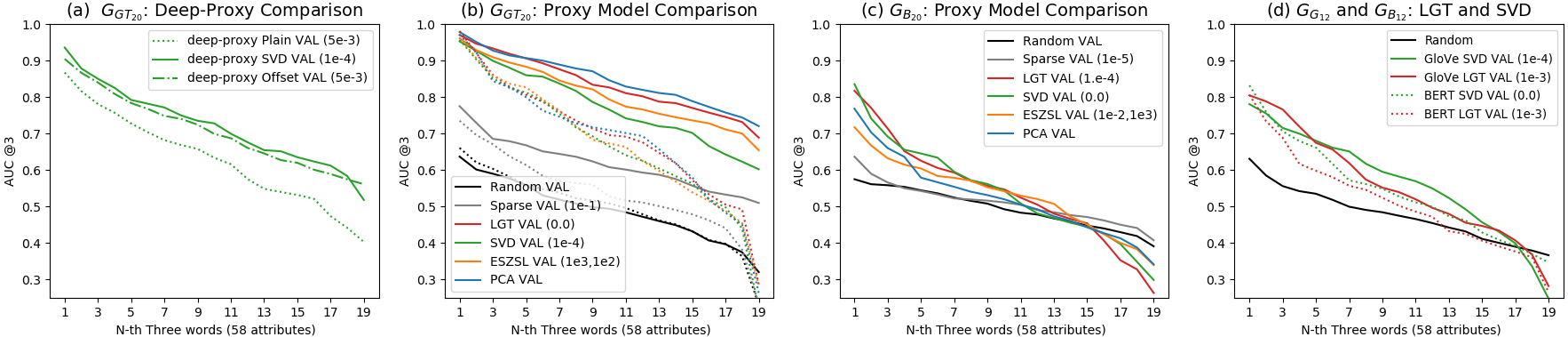}
  \caption{(a) Three deep-proxy models by $G_{GT_{20}}$ are compared on ``eval''. SVD is selected as the best model for art. (b) Proxy models by $G_{GT_{20}}$  are compared. The solid-lines are evaluated by ``eval-$G$'', and the dotted-lines are evaluated by ``eval-$NG$''. (c) Five proxy models by $G_{B_{20}}$ are evaluated by ``eval''. (d) SVD and LGT by $G_{B_{12}}$ and  $G_{G_{12}}$ are compared by ``eval-VAL''.}
\label{fig:fig_3}
\end{figure*}
\subsubsection{Model Selection for Deep-Proxy}
To select the best deep-proxy for art, the three  versions of Plain, SVD, and Offset by  $G_{GT_{20}}$ are compared.  For Offset,  the $G_{GT_{20}}$ is pre-shifted by $+1.0$ to make all components of $G_{GT_{20}}$ matrix  positive,  and let machines learn a new offset  from  it.  For the regularization  $\lambda$ in  equation (\ref{eqn:eqn_4}), $1\mathrm{e}{-4}$, $5\mathrm{e}{-4}$, $1\mathrm{e}{-3}$, and $5\mathrm{e}{-3}$ are tested. In Figure \ref{fig:fig_3} (a), SVD  achieved the best rates and  outperformed the Plain model. Offset  was not  as effective as SVD. Since  $G_{GT_{20}}$ was computed from the ground truth values, its zero point was originally aligned with ``somewhat'', so offset learning may not be needed. 
\par For a comparable result with SUN data, Offset is shown as the best  in Figure \ref{fig:fig_4} (a).  SUN's $G$ matrix is computed by ``binary'' ground truths, so it is impossible to gauge the right offset. Hence, Offset learning becomes advantageous for SUN. However, for CUB, SVD and Offset were not learnable (converged to a local minimum, whose recognition is the random choice of  equal probabilities). Since  CUB's $G$ matrix has  smaller eigenvalues than other data sets, implying  subtle visual difference among bird classes, the two deep-proxy methods happen to be infeasible by demanding a neural net to capture  fine and local visual differences of birds first,  in order to  discern  the birds as orthogonal vectors. 
However, for the neural net especially in the initial stage of  learning,  finding the right direction to the high goal is far more challenging compared  to the case of art and SUN, whose attributes can be found rather distinctively and globally in different class images. The detailed results for CUB and SUN are shown in SI E.

\subsubsection{Sensitive Methods to Language Models} 
\par Proxy models by $G_{GT_{20}}$ and $G_{B_{20}}$ are evaluated in Figure \ref{fig:fig_3} (b) and (c). To avoid the bias by the samples used in computing $G$ matrix, for the models by $G_{GT_{20}}$, validation (solid-line) and test (dotted-line) are separately computed based on ``eval-$G$'' and ``eval-$NG$ '' each.
\par High sensitivity to $G_{B_{20}}$ is observed for PCA and ESZSL. In Figure \ref{fig:fig_3} (b),  PCA  and LGT show similar performance  on ``eval-$NG$'', but on ``eval-$G$'', PCA performs better than LGT. The same phenomenon  is observed between ESZSL and SVD again. The better performance on ``eval-$G$'' indicates somewhat direct replication of $G$ matrix into  outcomes. This hints that ESZSL and PCA can suffer more  degradation than other models if $G$ matrix is estimated by language models, so its imperfection straightly  act on their results, as shown in Figure \ref{fig:fig_3} (b) and (c). Since they compute visual elements through direct multiplications between processed features and $G$ matrix, and particularly for ESZSL, it finds a global optimum  given a $G$ matrix,  they showed the highest sensitivity to the condition of $G$ matrix.

\subsubsection{Robust Methods to Language Models} 
Deep-learning makes LGT and deep-proxy slowly adapt to the information given by $G$ matrix, so the models are less affected by language models than ESZSL and PCA, as shown in Figure \ref{fig:fig_3} (c). LGT can learn some visual elements through BERT or GloVe, even when not all style relations for the elements are correct in the models. For  example, for $G_{B}$, `expressionism' is encoded as more related with ``abstract'' than 'cubism' or `abstract-expressionism', which is false. But despite the partial incorrectness, LGT  gradually learns the semantic of ``abstract'' at the rate of $0.84$ AUC using the training data in a larger range of styles, correctly encoded; northern-renaissance (least related to ``abstract'') $<$ rococo $<$ cubism $<$ abstract-expressionism (most related to ``abstract'') etc (abstract AUCs of SVD, PCA, and ESZSL by $G_{B_{20}}$: $0.9, 0.8, 0.7$).
\par Deep-proxy  more actively adjusts some portion of $G$ matrix. Suppose there is a neural net trained with $G' = (G+\Delta G)$ distorted by $\Delta G$. By equation (\ref{eqn:eqn_2}), this $f_{A}'(x)^{t} \cdot (G+\Delta G)=f_{S}^{t}(x)$ is a valid convergent point of the net, and we also can see this $(f_{A}'(x)^{t} + b^{t}(x)) \cdot G=f_{S}(x)^{t}$ as another possible point, where $b(x)^{t} \cdot G = f_{A}'(x)^{t} \cdot \Delta G$. If the bottom  of the neural net approximately learns $f_{A}'(x)^{t} + b^{t}(x)$, it would work as if a better $G$ is given, absorbing some errors. This adjustment could explain the robustness to the imperfection of language models than others, and also the flexibility to the sparse $G$ matrix that is shown to be directly harmful for LGT.  This will be discussed in the next section.
\begin{figure}[h]
 \centering
 \includegraphics[width=0.8\linewidth]{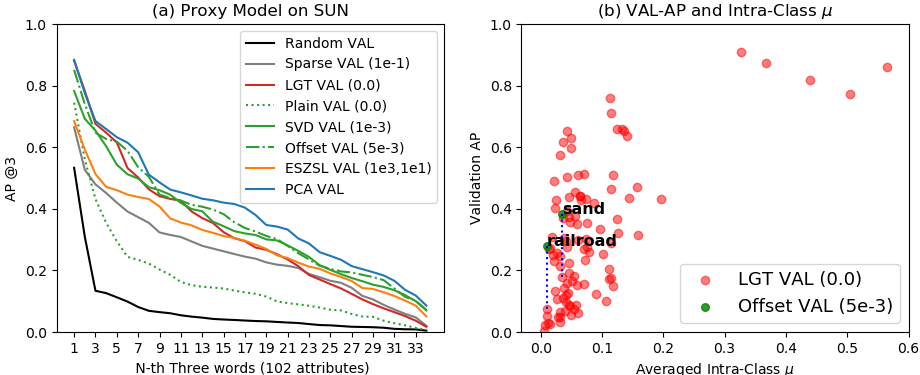}
  \caption{SUN experiment: (a) Validation results for all  models are shown. (b)  The relationship between validation-AP (y-axis) and intra-class $\mu$ (x-axis) for LGT is presented for each  attribute (each red dot). Points of Offset (two green dots) are drawn only when the AP-gap between  Offset and LGT is more than $0.2$.  The higher scores on the green dots show that Offset is less affected by the sparsity of $G$ matrix than LGT.  As the AP-gap gets lower to $0.15$, $19$ words were found, and Offset worked better than LGT for all the $19$ words.}
\label{fig:fig_4}
\end{figure}
\begin{figure*}[t!]
 \centering
\includegraphics[width=1.0\linewidth]{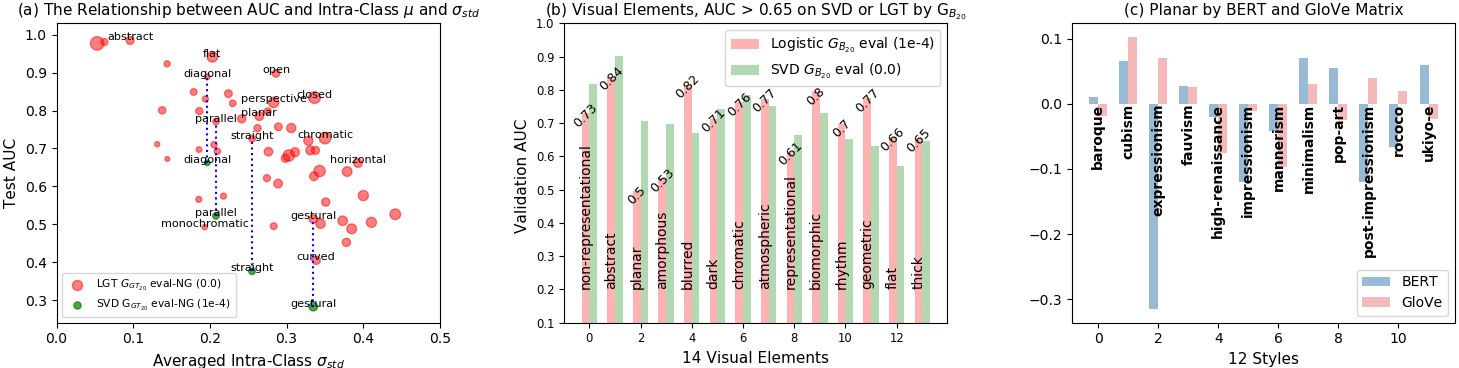}
  \caption{(a)  The relationship between test-AUC (y-axis) and  $\sigma_{std}$ (x-axis) and  $\mu$ (size of dots) for LGT (by $G_{GT_{20}}$ and ``eval-NG'') is shown for each attribute (each red dot). Offset points (four green dots) are drawn only when the AUC-gap between the SVD and LGT is more than $0.2$. Each performance gap is traced by the blue dotted-lines.  (b) Visual elements scored more than $0.65$ AUC by SVD or LGT (by $G_{B_{20}}$ and ``eval'') are presented. (c) Style Encoding of BERT and GloVe for the word ``planar''.}
\label{fig:fig_5}
\end{figure*}
\subsubsection{Logistic and Deep-Proxy on $G_{GT}$} Two factors are analyzed  with LGT and SVD  performance: intra-class’s standard deviation ($\sigma_{std}$) and mean ($\mu$). The intra-statistics of each style are computed with ``eval'' and  averaged across the styles to estimate $\sigma_{std}$  and $\mu$  for $58$ visual elements. For LGT and SVD by $G_{GT_{20}}$, AUC is moderately related with $\sigma_{std}$ (Pearson correlation coefficient  \textit{$r_{LGT}$} = $-0.65$ and \textit{$r_{SVD}$} = $-0.51$), but their performance is not explained solely by $\sigma_{std}$. As shown in Figure \ref{fig:fig_5} (a), ``monochromatic'' (AUC of LGT and SVD: $0.49$ and $0.58$) scored far less than ``flat'' (AUC of LGT and SVD : $0.94$ and $0.92$) even if both words' $\sigma_{std}$ are similar and small. Since the element of monochromatic was not a relevant feature for most of styles, it was consistently encoded as very small values across the styles in $G_{GT}$ matrix. The element has small variance within a style, but does not have enough power to discriminate styles so failed to learn. LGT can be  more degraded with the sparsity because the information encoded  that is close to zero for all styles cannot be back-propagated properly. As shown in Figure \ref{fig:fig_4} (b), intra-class $\mu$ of $102$ attributes in SUN are densely distributed  between $0.0$ and $0.1$, so LGT is lower ranked  compared to art. LGT AP is most tied in the sparse $\mu$ to others ($r_{LGT}$ = $0.43$,  $r_{Offset}$ = $0.36$, $r_{PCA}$ = $0.33$,  $r_{ESZSL}$ = $-0.15$ at $\mu < 0.3$).
\par For SVD by $G_{GT_{20}}$, its overall performance is   lower than LGT by  $G_{GT_{20}}$. When the words ``diagonal'', ``parallel'', ``straight'', and ``gestural'' (four green dots in Figure \ref{fig:fig_5} (a)) were found by the condition of $|AUC_{(SVD)}-AUC_{(LGT)}|>0.2$, LGT scored higher than SVD  for all words. Since SVD is trained by an objective function for multiple style classification, the learning visual elements can be restricted by the following cases. Some hidden axes could be used to learn  non-semantic features  to promote learning styles. 
Or, some necessary semantics for styles could be strewn throughout multiple axes. Hence, LGT generally learns more words than SVD when $G$ matrix is estimated by some ground truths as shown in Figure \ref{fig:fig_3} (b), but $G$ matrix should not be too sparse for LGT.

\begin{figure}[h!]
 \centering
  \includegraphics[width=0.65\textwidth]{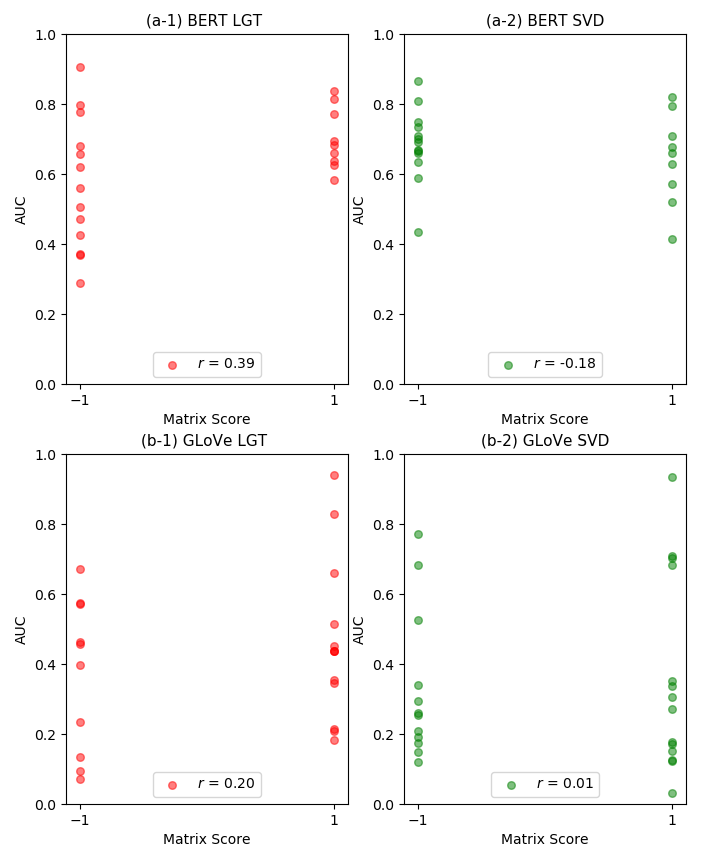}
  \caption{Correlation analysis between AUCs and matrix scores. The BERT plots of (a-1) and (a-2) are drawn based on the $22$ visual elements which scored more than AUC $0.6$ by any of SVD or LGT (by $G_{B_{12}}$). The GloVe plots of (b-1) and (b-2) are drawn based on the $28$ visual elements which scored more than AUC $0.6$ by any of SVD or LGT (by $G_{G_{12}}$). This shows LGT is more sensitively affected by the quality of language models.
  }
 \label{fig:fig_6}
\end{figure}
\begin{figure}[h]
 \centering
  \includegraphics[width=0.8\textwidth]{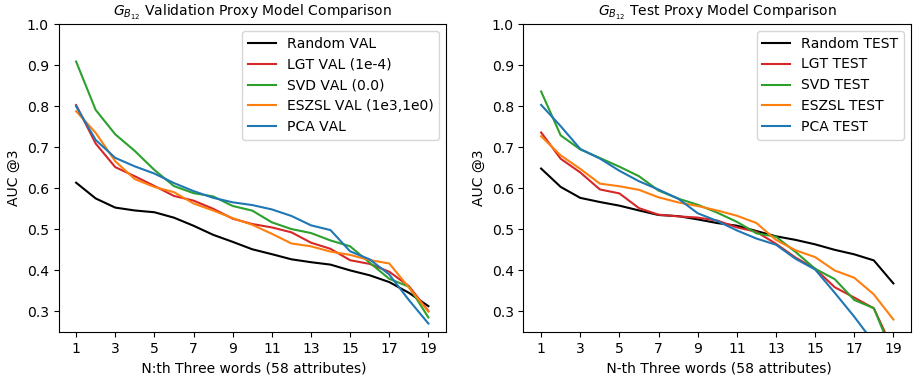}
  \caption{AUC plots of random split on BERT: SVD by $G_{B_{12}}$ scored better than LGT at most of AUC@$3$ points.}
 \label{fig:fig_7}
\end{figure}
\subsubsection{Logistic and Deep-Proxy on BERT and GloVe}
For  language models, it is a bit hard to generalize  the performance of LGT and SVD. As shown in Figure \ref{fig:fig_5} (b), it was not clear which is better with BERT. We needed another comparable language model to understand their performance. GloVe is tested after dividing $20$ styles  into train ($12$ styles) \footnote{Baroque, Cubism, Expressionism, Fauvism, High-Renaissance, Impressionism, Mannerism, Minimalism, Pop-Art, and Ukiyo-e} and test ($8$ styles). Aligned with the split, the ``eval'' was also separated into ``eval-VAL'' and ``eval-TEST '' ($8$ unseen styles in training). Here, the ``eval-VAL'' was used to select hyper parameters.  On the same split, the models by BERT $G_{B_{12}}$ were compared, too. Depending on each language model, the ranking relations were differently shown. In  Figure \ref{fig:fig_3} (d),  SVD by  $G_{B_{12}}$ scored better than LGT at all AUC@$3$ points. However, LGT by $G_{G_{12}}$  was better than SVD for the first top $15$  words, but for second  $15$ words, SVD scored better than LGT. To figure out a key factor of the different performance,  we scored the quality of BERT and GloVe with $\{-1,+1\}$ for each visual element and conducted correlation analysis between the scores and the AUC results. Pearson correlation coefficient \textit{$r$} between AUCs and the scores are computed. The results are shown in Figure   \ref{fig:fig_6}.
\par In the analysis, GloVe scored higher than BERT, and  LGT  showed the stronger correlation than SVD between AUCs and scores. This proves the robustness of SVD to the imperfection of language models along with  the results of Fig \ref{fig:fig_3} (d).
As a specific example,  the word  ``planar'' is incorrectly encoded by BERT, quantifying some  negatives on Expressionism, Impressionism, and Post-Impressionism as shown by Figure \ref{fig:fig_5} (c). The wrong information influenced more on LGT, so its AUC scored $0.38$ (eval-TEST: $0.47$)  by BERT  but   $0.77$ (eval-TEST: $0.76$)   by GloVe, while SVD learned ``planar'' by the similar rates of  $0.73$  (eval-TEST: $0.58$) by BERT  and $0.78$  (eval-TEST: $0.68$) by GloVe on ``eval-VAL''. For LGT, the defective information is directly provided through training data, so it is more sensitively affected by noisy language models. On the other hand, SVD can learn some elements even when it is trained by a $G$ matrix that is not perfect if the elements are  essential for  style classification possibly through the adjustment operation, as aforementioned. Another  split  of 12-training style \footnote{Art-Nouveau-Modern, Color-Field-Painting,  Early-Renaissance, Fauvism, High-Renaissance, Impressionism, Mannerism, Northern-Renaissance, Pop-Art, Rococo, Romanticism, and Ukiyo-e} vs. 8-test style is also tested by BERT. In this experiment, LGT also scored less than SVD as shown in Figure \ref{fig:fig_7}.
\begin{table}[t!]
\centering
\resizebox{1.0\columnwidth}{!}{
\begin{tabular}{ccccccc} 
\toprule
\multicolumn{5}{c}{Visual Elements (AUC $\geq$ 0.65) }&\multicolumn{2}{c}{Visual Elements (AUC $\leq$ 0.65)}\\
\cmidrule(r){1-5}
\cmidrule(r){6-7}
abstract&chromatic&atmospheric&planar&representational&geometric&perspective\\
\midrule
\includegraphics[height=1.6in,width=1.6in]{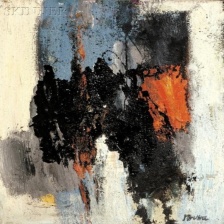}&
\includegraphics[height=1.6in,width=1.6in]{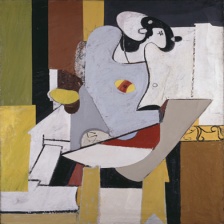}&
\includegraphics[height=1.6in,width=1.6in]{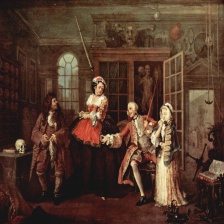}&
\includegraphics[height=1.6in,width=1.6in]{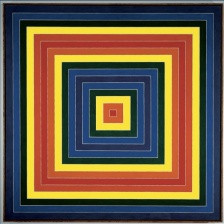}&
\includegraphics[height=1.6in,width=1.6in]{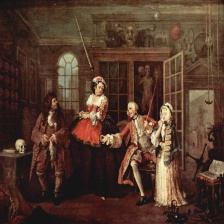}&
\includegraphics[height=1.6in,width=1.6in]{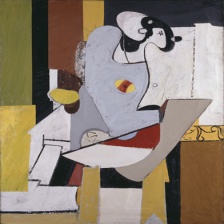}&
\includegraphics[height=1.6in,width=1.6in]{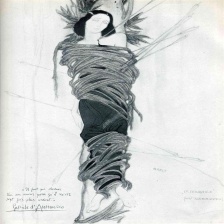}\\
\includegraphics[height=1.6in,width=1.6in]{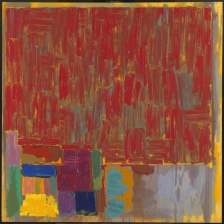}&
\includegraphics[height=1.6in,width=1.6in]{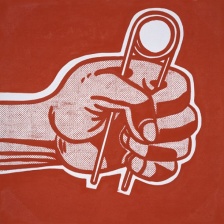}&
\includegraphics[height=1.6in,width=1.6in]{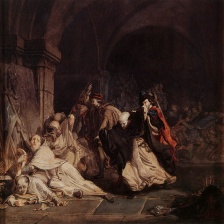}&
\includegraphics[height=1.6in,width=1.6in]{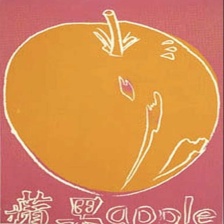}&
\includegraphics[height=1.6in,width=1.6in]{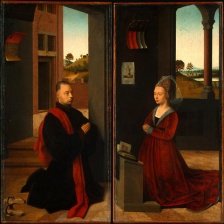}&
\includegraphics[height=1.6in,width=1.6in]{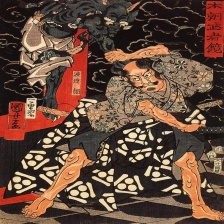}&
\includegraphics[height=1.6in,width=1.6in]{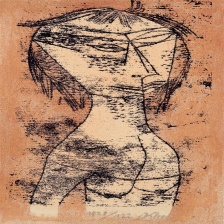}\\

\includegraphics[height=1.6in,width=1.6in]{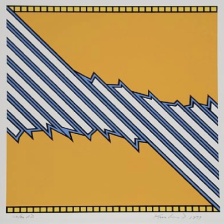}&
\includegraphics[height=1.6in,width=1.6in]{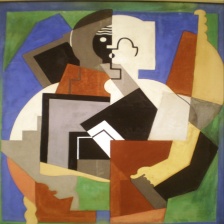}&
\includegraphics[height=1.6in,width=1.6in]{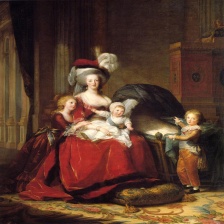}&
\includegraphics[height=1.6in,width=1.6in]{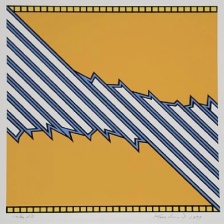}&
\includegraphics[height=1.6in,width=1.6in]{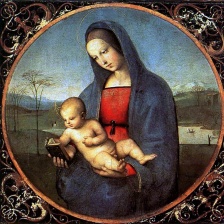}&
\includegraphics[height=1.6in,width=1.6in]{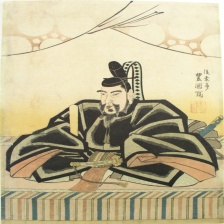}&
\includegraphics[height=1.6in,width=1.6in]{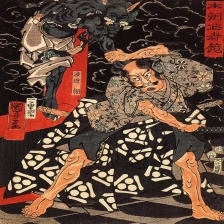}\\

\includegraphics[height=1.6in,width=1.6in]{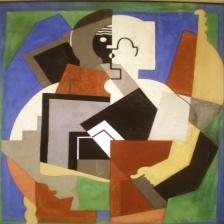}&
\includegraphics[height=1.6in,width=1.6in]{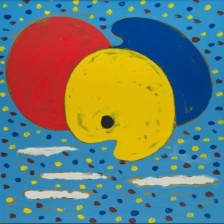}&
\includegraphics[height=1.6in,width=1.6in]{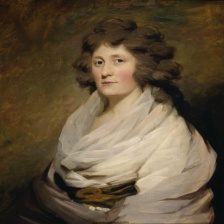}&
\includegraphics[height=1.6in,width=1.6in]{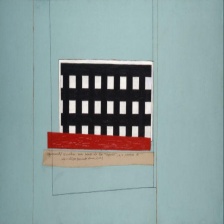}&
\includegraphics[height=1.6in,width=1.6in]{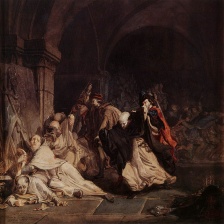}&
\includegraphics[height=1.6in,width=1.6in]{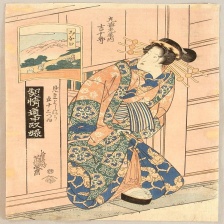}&
\includegraphics[height=1.6in,width=1.6in]{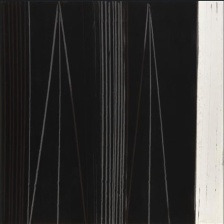}\\

\includegraphics[height=1.6in,width=1.6in]{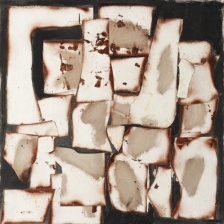}&
\includegraphics[height=1.6in,width=1.6in]{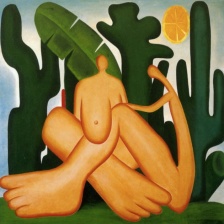}&
\includegraphics[height=1.6in,width=1.6in]{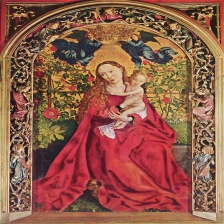}&
\includegraphics[height=1.6in,width=1.6in]{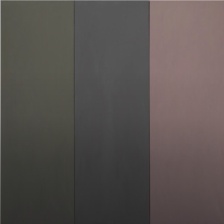}&
\includegraphics[height=1.6in,width=1.6in]{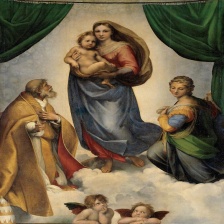}&
\includegraphics[height=1.6in,width=1.6in]{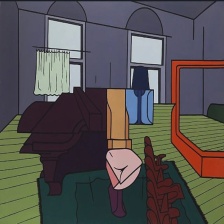}&
\includegraphics[height=1.6in,width=1.6in]{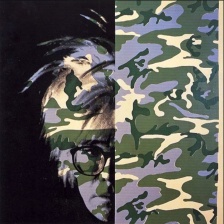}\\\hline\\\hline

\includegraphics[height=1.6in,width=1.6in]{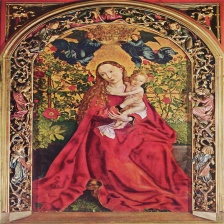}&
\includegraphics[height=1.6in,width=1.6in]{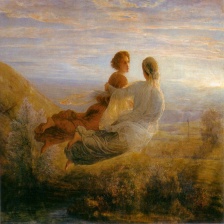}&
\includegraphics[height=1.6in,width=1.6in]{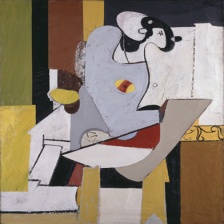}&
\includegraphics[height=1.6in,width=1.6in]{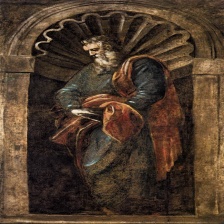}&
\includegraphics[height=1.6in,width=1.6in]{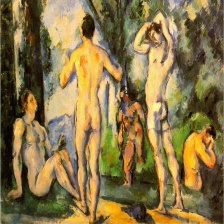}&
\includegraphics[height=1.6in,width=1.6in]{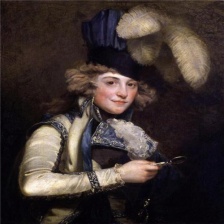}&
\includegraphics[height=1.6in,width=1.6in]{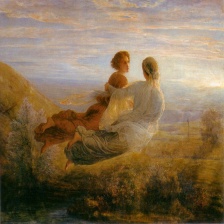}\\

\includegraphics[height=1.6in,width=1.6in]{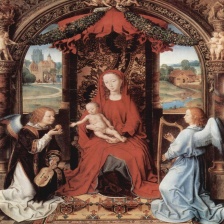}&
\includegraphics[height=1.6in,width=1.6in]{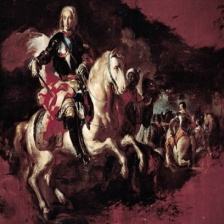}&
\includegraphics[height=1.6in,width=1.6in]{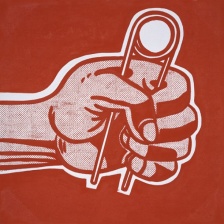}&
\includegraphics[height=1.6in,width=1.6in]{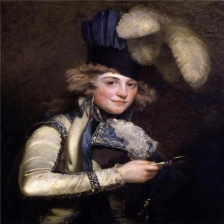}&
\includegraphics[height=1.6in,width=1.6in]{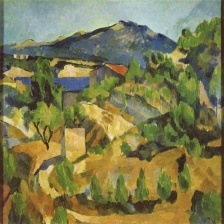}&
\includegraphics[height=1.6in,width=1.6in]{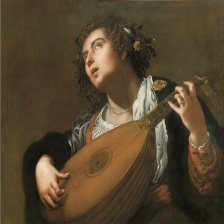}&
\includegraphics[height=1.6in,width=1.6in]{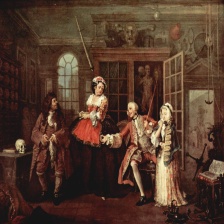}\\

\includegraphics[height=1.6in,width=1.6in]{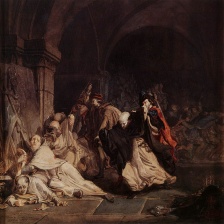}&
\includegraphics[height=1.6in,width=1.6in]{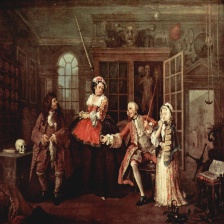}&
\includegraphics[height=1.6in,width=1.6in]{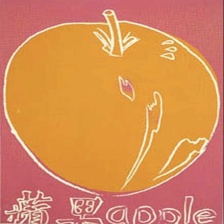}&
\includegraphics[height=1.6in,width=1.6in]{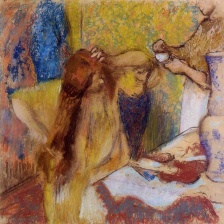}&
\includegraphics[height=1.6in,width=1.6in]{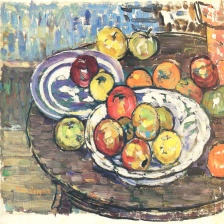}&
\includegraphics[height=1.6in,width=1.6in]{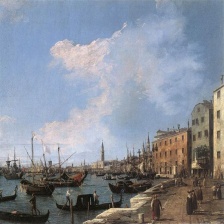}&
\includegraphics[height=1.6in,width=1.6in]{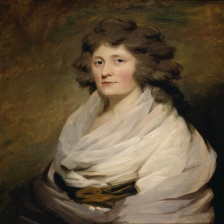}\\

\includegraphics[height=1.6in,width=1.6in]{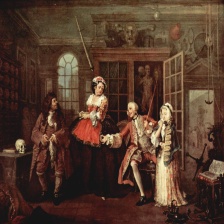}&
\includegraphics[height=1.6in,width=1.6in]{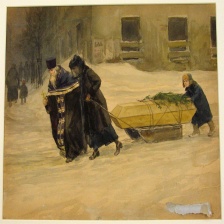}&
\includegraphics[height=1.6in,width=1.6in]{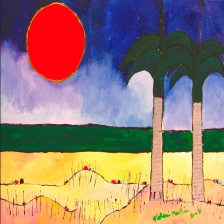}&
\includegraphics[height=1.6in,width=1.6in]{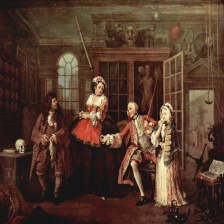}&
\includegraphics[height=1.6in,width=1.6in]{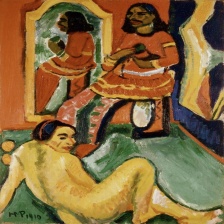}&
\includegraphics[height=1.6in,width=1.6in]{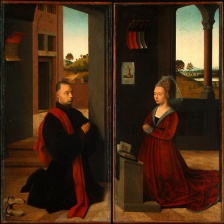}&
\includegraphics[height=1.6in,width=1.6in]{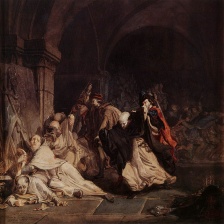}\\

\includegraphics[height=1.6in,width=1.6in]{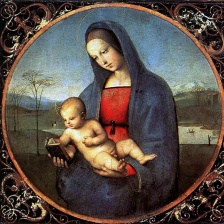}&
\includegraphics[height=1.6in,width=1.6in]{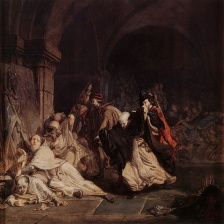}&
\includegraphics[height=1.6in,width=1.6in]{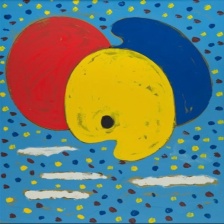}&
\includegraphics[height=1.6in,width=1.6in]{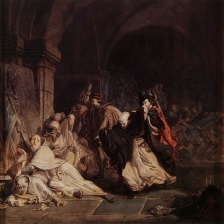}&
\includegraphics[height=1.6in,width=1.6in]{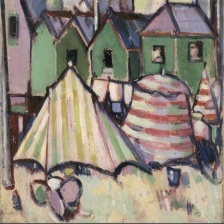}&
\includegraphics[height=1.6in,width=1.6in]{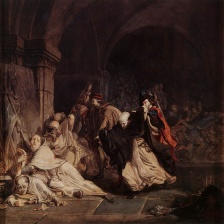}&
\includegraphics[height=1.6in,width=1.6in]{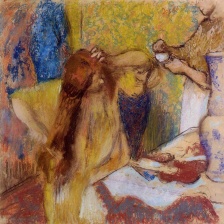}\\\hline
0.90&0.79&0.75&0.71&0.67&0.63&0.46\\\hline
\end{tabular}}

\caption{Descending ranking results (top to bottom) based on the prediction $f_{A}(x)$ of SVD ($G_{B_{20}}$  and $\lambda = 0.0$). The five most ($1-5$ rows) and five least ($6-10$ rows) relevant paintings are shown as the machine predicted. The last row indicates the AUC score of each visual element.} 
\label{tab:tab_1}
\end{table}
\subsubsection{Descending Ranking Results of SVD by $G_{B_{20}}$}
To present some example results, 120 paintings of ``eval'' are sorted based on the activation values $f_{A} (x)$ of SVD by $G_{B_{20}}$. Table \ref{tab:tab_1} presents some  results of words that achieved more than $0.65$ or less than $0.65$ with BERT model. This shows how the ``eval'' paintings are visually different according to each output-value of deep-proxy-$G_{B}$ for the selected seven visual elements (abstract, chromatic, atmospheric, planar, representational, geometric, and perspective). 

\section{Conclusion and Future Work}
\par Quantifying fine art paintings based on visual elements is a fundamental part of  developing AI systems for art, but their direct annotations are very scarce. In this paper, we presented several proxy methods to learn the valuable information  through its general and linear relations to style, which can be estimated by language models or human survey. They are quantitatively analyzed to reveal  how the inherent structures of the methods make them robust or weak on the practical estimation scenarios. The robustness of  deep-proxy to the imperfection of language models is a key finding. For future study,  we will look at more complex systems. For  example, a non-linear relation block learned by language models could be transferred or transplanted to a neural network to learn visual elements through the deeper relation with styles.  Furthermore,  direct applications for finding acoustic semantics for music genres or learning principle elements for fashion designs would be interesting subjects for proxy learning. Their attributes are visually or acoustically shared to define higher level of categories, but their class boundaries could be softened as proxy representations.
\clearpage
\bibliographystyle{unsrt}  
\bibliography{references}
\clearpage
\appendix
\renewcommand\thesection{\Alph{section}}
\renewcommand\thesubsection{\thesection.\arabic{subsection}}
\counterwithin{figure}{section}
\counterwithin{table}{section}
\section*{Supplementary Information (SI)}

\section{Painting Information}
\begin{table*}[h!]
\centering
\resizebox{1.0\textwidth}{!}{

  \begin{tabular}{m{1.0 in}||c|m{1.0 in}||c}
  \hline
\multirow{3}{*}{}&&&\\
Painting&Information&Painting&Information\\
&&&\\\hline
\multirow{8}{*}{\includegraphics[height=1.0in, width=1.0in]{Images/Table_Results/97.jpg}}&&\multirow{8}{*}{\includegraphics[height=1.0in, width=1.0in]{Images/Table_Results/28.jpg}}&\\
&&&\\
&\makecell{title: Madonna Conestabile}&&\makecell{title:  The Architect, Jesus T. Acevedo}\\
&\makecell{author: Raphael}&&\makecell{author: Diego Rivera}\\
&\makecell{year: 1502}&&\makecell{year: 1915}\\
&\makecell{style: High Renaissance}&&\makecell{style: Cubism}\\
&&&\\
&&&\\\hline

\multirow{8}{*}{\includegraphics[height=1.0in, width=1.0in]{Images/Table_Results/98.jpg}}&&\multirow{8}{*}{\includegraphics[height=1.0in, width=1.0in]{Images/Table_Results/1.jpg}}&\\
&&&\\
&\makecell{title: The Sistine Madonna}&&\makecell{title: Water of the Flowery Mill}\\
&\makecell{author: Raphael}&&\makecell{author: Arshile Gorky}\\
&\makecell{year: 1513}&&\makecell{year: 1944}\\
&\makecell{style: High Renaissance}&&\makecell{style: Abstract Expressionism}\\
&&&\\
&&&\\\hline

\multirow{8}{*}{\includegraphics[height=1.0in, width=1.0in]{Images/Table_Results/99.jpg}}&&\multirow{8}{*}{\includegraphics[height=1.0in, width=1.0in]{Images/Table_Results/68.jpg}}&\\
&&&\\
&\makecell{title: Judith}&&\makecell{title: Pendulum}\\
&\makecell{author: Gorreggio}&&\makecell{author: Helen Frankenthaler}\\
&\makecell{year: 1512-1514}&&\makecell{year: 1972}\\
&\makecell{style: High Renaissance}&&\makecell{style: Abstract}\\
&&&\\
&&&\\\hline

\multirow{8}{*}{\includegraphics[height=1.0in, width=1.0in]{Images/Table_Results/79.jpg}}&&\multirow{8}{*}{\includegraphics[height=1.0in, width=1.0in]{Images/Table_Results/108.jpg}}&\\
&&&\\
&\makecell{title: Morning in a Village}&&\makecell{title: Untitled Vessel \#10}\\
&\makecell{author: Fyodor Vasilyev}&&\makecell{author: Denise Green}\\
&\makecell{year: 1869}&&\makecell{year: 1977}\\
&\makecell{style: Realism}&&\makecell{style: Neo-Expressionism}\\
&&&\\
&&&\\\hline

\multirow{8}{*}{\includegraphics[height=1.0in, width=1.0in]{Images/Table_Results/19.jpg}}&&\multirow{8}{*}{\includegraphics[height=1.0in, width=1.0in]{Images/Table_Results/9.jpg}}&\\
&&&\\
&\makecell{title: Forest Landscape with Stream}&&\makecell{title: Untitled No. 40 }\\
&\makecell{author: Ivan Shishkin}&&\makecell{author: Claude Viallat}\\
&\makecell{year: 1870-1880}&&\makecell{year: 1996}\\
&\makecell{style: Realism}&&\makecell{style: Color-Field-Painting}\\
&&&\\
&&&\\\hline\hline
\end{tabular}}

\caption{Information of title, author, year of made, and style}
\end{table*}

\section{BERT Model}
\subsection{Training BERT}
\begin{table}[h]
\centering
\begin{tabular}{l} 
\toprule
\multicolumn{1}{c}{List of Words}  \\
\midrule
non-representational, representational, meandering, gestural\\
amorphous, biomorphic, planar, chromatic\\
monochromatic, bumpy, art-nouveau, cubism\\
expressionism, fauvism, abstract-expressionism\\ color-field-painting, minimalism, naive-art, ukiyo-e\\
early-renaissance, pop-art, high-renaissance, mannerism\\
northern-renaissance, rococo, romanticism, impressionism\\
post-impressionism\\
\bottomrule
\end{tabular}
\label{tab:tabA_1}
\caption{28 words newly added to the original dictionary of BERT}
\end{table}
For a new BERT model for art, the BERT-BASE model (12-layer, 768-hidden, 12-heads, and not using cased letters) was selected and trained from scratch over the collected art corpus. For  training, the original BERT vocabulary is updated by adding new words. Table B.1 shows the words that are newly added.  For optimizer, \texttt{Adam algorithm with decay} is used. The BERT model is trained for  $5\mathrm{e}{+5}$ steps with the \texttt{learning rate} of $1.0\mathrm{e}{-4}$ and  the \texttt{number of warm-up steps} is set to  $1\mathrm{e}{+4}$.

\subsection{Collecting BERT Embedding}
Context-free embedding is collected for each of art terms (20 styles and 58 visual elements). The context-free means only a target word is inputted to BERT to collect word-embedding without accompanying other words.  Each target word is enclosed only by [CLS] and [SEP] and inputted to the BERT  as the format of ``\texttt{[CLS] target-word [SEP]}''.  12 output vectors (768-D) from all 12 hidden layers are collected and averaged. The representations corresponding to the [CLS] and [SEP] are discarded so only the vector representations for input words are taken as the final embedding. We also tried to average the embeddings collected only from the  top  or bottom four  layers, but all 12 embeddings  were  slightly better at presenting the general relationship between styles and visual elements. 
\begin{figure}[h!]
  \centering
  \includegraphics[width=\textwidth]{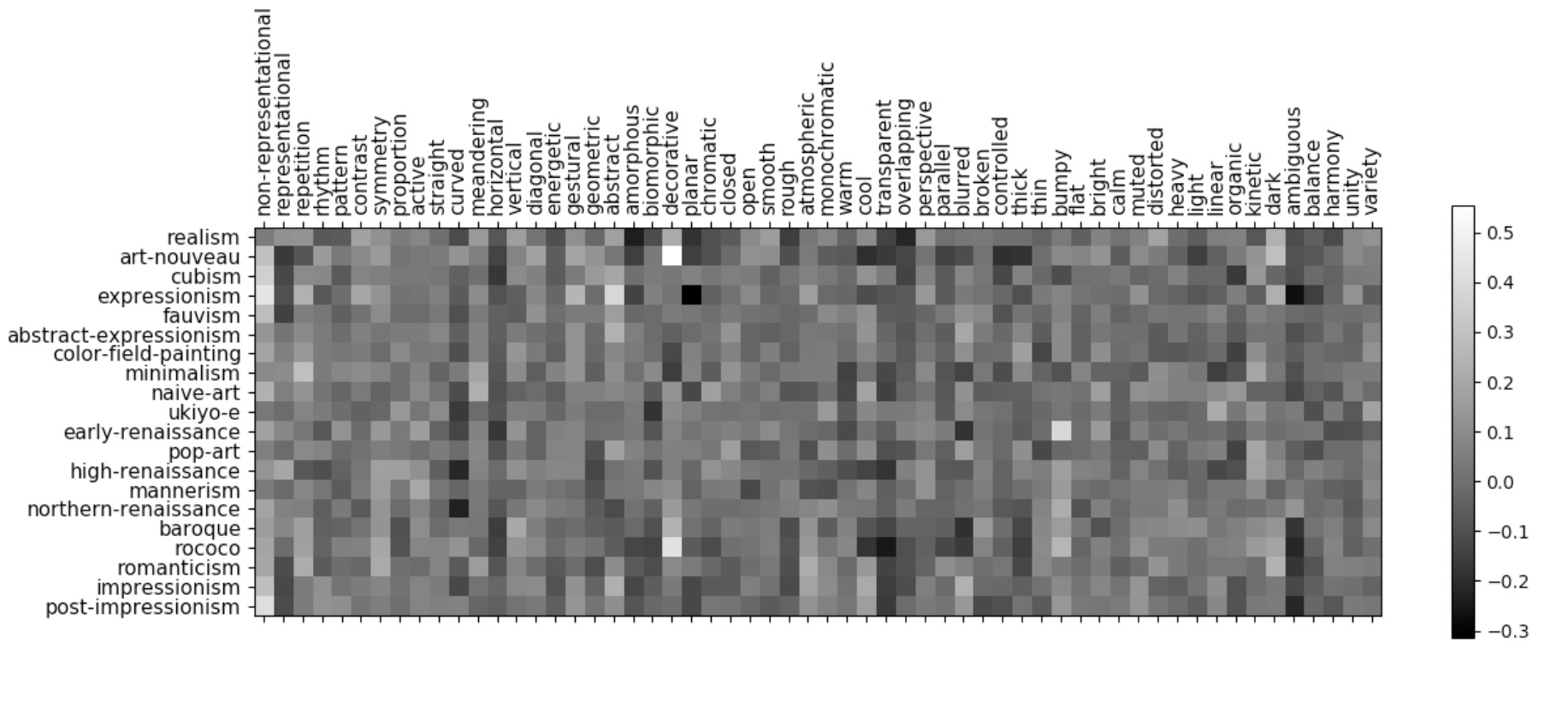}
  \caption{A general relation between 58 visual elements and 20 styles is estimated by BERT.  It is visualized in gray-scale.}
  \label{fig:figA_1}
\end{figure}

\subsection{Visualization of BERT matrix}
The general relation estimated by BERT is visualised in gray-scale in Figure \ref{fig:figA_1}.  In this figure,  the birther square shows the stronger association between  styles and  visual elements. 
\section{Four Proxy Models}
\subsection{Sparse Coding}
Sparse coding \cite{efron2004least} is the simplest model for proxy learning, which estimates the visual information from style encodings  without seeing input paintings. By treating  the rows of $G$ matrix as over-complete bases of style space  $\mathbf{S}$ (the number of styles in this paper is at most 20 and the number of visual elements is 58),    $i_{a}$ ($f_A(x)$) is estimated  from  given $G$ and $f_S(x)$ by sparse coding equation  (\ref{eqn:eqnB_1}) below. 
\par 
\begin{equation}
\argmin_{i_{a}}\lVert{f_S(x)-G^{t}\cdot i_{a}}\rVert_{2} + \lambda_{s} \cdot \lVert{i_{a}}\rVert_{1}
\label{eqn:eqnB_1}
\end{equation}
\par To encode a painting $x$ into mutliple styles ($f_S(x)$), the soft-max output of a style classifier is used. To implement a style classifier, VGG-16 ImageNet's \cite{Simonyan15} all convolution layers are transferred as is and frozen, but its original FC layers $(4096 - 4096 - 1000)$ are modified to five FC layers $(4096 -  4096  - 1024 - 512 - n \text{ number of styles})$ and their parameters are updated in training.   The hyper-parameter $\lambda_{s}$ tried during  development  are $0.0$, $1\mathrm{e}{-1}$, $1\mathrm{e}{-2}, 1\mathrm{e}{-3}, 1\mathrm{e}{-4} ,1\mathrm{e}{-5}, 0.15, 0.3, 0.45$.  The linear equation (\ref{eqn:eqnB_1}) is solved by Least Angle Regression \cite{efron2004least}.
\subsection{Logistic Regression}
Logistic regression \cite{danaci2016low} (LGT) is used to learn visual elements in a sort of supervised way.  Similar to the work of Lampert et al. \cite{lampert2013attribute}, each column of $G$ was used to assign attributes to images on a per class basis. Each column index was determined based on the style label. When $G$ matrix  are not probabilistic representations, without shifting zero points, the positives were put into the range of $0.5$ to $1.0$ and the negatives were put into the range of $0.0$ to $0.5$.  Finally, the values were adjusted from zero to one. 
\par The logistic regression is implemented on the last convolution layer of a VGG-16 ImageNet. After adding multiple FC layers $(2048 - 2048 - 1024 - 58)$ on the top of the convolution layer, same as  deep-proxy, and the FC parts are newly updated by an objective function. Let  $I(q,k)$ be an indicator function that states whether or not the $k$-th sample belongs to style class $q$. Let $\vec{g_{q}}$ (58-D) be the probabilistic representation of a column of G matrix corresponding to $q$ style class. Let  $\vec{l}(x|\omega)$ be a (58-D) logistic output of the network, where $x$ is an input image, $\omega$ implies the network's parameter, and $\omega_{1024 \times 58}$ denotes the last FC layer. Then, an objective function for logistic regression is set as in equation (\ref{eqn:eqnB_2}) below. 
\par 
\begin{equation}
\min_{\omega} \sum_{k}^{K} \sum_{q}^{Q} - I(q,k) \cdot \Big\{ \vec{g_{q}}^{t} \cdot \log_{e}( \vec{l}(x|\omega)) + (\vec{1}_{58} - \vec{g_{q}})^{t}
\cdot  \log_{e}(\vec{1}_{58} -\vec{l}(x|\omega)) \Big\} + \lambda_{L} \cdot \lVert{\omega_{1024 \times 58}}\rVert_{1} 
\label{eqn:eqnB_2}
\end{equation}
\par The  $\lambda_{L}$ (regularization) is added to   reduce  undesirable correlations among  attributes by restricting the magnitudes of the last layer parameters. Tested $\lambda_{L}$s are $0.0$, $1\mathrm{e}{-3}$, $1\mathrm{e}{-4}$, and $1\mathrm{e}{-5}$. All logistic models are trained for 200,000 steps at 32 batch size by a momentum optimizer. Their learning rates are initially set as $1\mathrm{e}{-3}$ and degraded at the factor of 0.94 every two epochs as same as  deep-proxy.
\subsection{PCA}
Kim et al. \cite{diana2018artprinciple}  proposed a framework based on Principal Component Analysis (PCA) to annotate paintings with a set  of visual concepts. Differently to    proxy learning, they consider a style conceptual space as a joint space where paintings and each visual concept can be properly encoded to measure their associations. Even though its approach to style space is different from proxy learning, PCA method can be reproduced in the context of proxy learning as follows.
\begin{enumerate}
    \item \textbf{PCA transform of visual embedding}  The visual embedding, collected from the last hidden layer (before activation) of a deep-CNN (style classification model), is transformed to PCA space.  A VGG-16 style classification model (FC part : $4096 - 4096 - 1024 -  512$) proposed by   Elgammal et al.  \cite{elgammal2018shape} is used to collect the embedding. Let us denote  $d$ is the dimension of the embedding,  $k$ is the number of painting samples,  and $p$ is the number of PCA components that cover 95\%  variance of the embedding. The last layer's  hidden embedding $E \in \mathbb{R}^{d \times k }$ is collected, and then  PCA projection matrix $P\in \mathbb{R}^{p \times d }$ (without whitening) is computed from $E$. Then, $E-[\vec{m}|\vec{m}|...|\vec{m}]$ is transformed to PCA samples $V \in\mathbb{R}^{p \times k} $  by equation  (\ref{eqn:eqnB_3}) below, where $\vec{m} \in\mathbb{R}^{d} $ is a sample mean of  $E$, and $\boldsymbol{m} = [\vec{m}|\vec{m}|...|\vec{m}]$ is the tiling matrix of  $\vec{m}$.
    \begin{equation}
    V = P \cdot (E-\boldsymbol{m})
    \label{eqn:eqnB_3}
    \end{equation}
    \item \textbf{Style encoding of PCA components} Let us denote each column of $H \in \mathbb{R} ^{n\times k}$ is one-hot binary vector to encode  a corresponding style label of each  sample in $V$, where  $n$ denotes number of styles. The multivariate linear regression equation (\ref{eqn:eqnB_4}) below is solved to compute $Z$.  In the solution $Z \in \mathbb{R} ^{p\times n}$,  each PCA-axis is encoded by $n$ styles.
    \begin{equation}
    V^{t} \cdot Z =  H^{t}
    \label{eqn:eqnB_4}
    \end{equation}
    \item \textbf{Computing visual attributes} Finally,  attribute representation $i_{a} \in \mathbb{R}^{58} (f_A(x))$  is computed by the equation (\ref{eqn:eqnB_5}) below, where $v \in  \mathbb{R}^{p}$ indicates a PCA representation of a test sample and the  $G \in \mathbb{R}^{58 \times n}$ is a category-attribute matrix. 
\begin{equation}
i_{a}^{t} = v^{t} \cdot Z \cdot G^{t}
\label{eqn:eqnB_5}
\end{equation}
\end{enumerate}
\subsection{ESZSL}
An Embarrassingly Simple Approach to Zero-shot Learning (ESZSL) \cite{romera2015embarrassingly} is  originally designed for zero-shot learning, not for learning visual semantics. However, its linear modeling between semantics and classes is comparable with deep-proxy, and   can be re-defined as a method for proxy learning as follows.
\begin{enumerate}
    \item \textbf{Computing $Q$ matrix to transform visual features to visual attributes} Let us denote $E\in \mathbb{R}^{d \times k}$ image features of $k$ training samples, where the $d$ is the dimension of the features. Let us denote $Y\in \{-1,1\}^{k \times n}$  the ground truth labels of each sample belonging to any of $n$ styles.  Then, the closed formulation (\ref{eqn:eqnB_6}) below  computes a single matrix $Q\in \mathbb{R}^{d \times 58}$. The single matrix $G$  is used to transform the image samples $E$ to  the visual elements. Two regularization parameters ($\lambda_{1e}$ and $\lambda_{2e}$) are used and all the possible combinations of $\lambda_{1e} = 10^{a}$ and $\lambda_{2e} = 10^{b}$ for $a,b = -3, -2, ..., 2, 3$ are considered. In this paper, the last hidden layer of VGG-16 ImageNet (4096-D before activation), generally called ``fc7-feature'', is used for the collection of  the image feature $E$,
\begin{equation}
Q =(E \cdot E^{t} + \lambda_{1e} \cdot  I)^{-1} \cdot E\cdot Y \cdot G^{t} \cdot (G \cdot G^{t} + \lambda_{2e} \cdot I)^{-1}
\label{eqn:eqnB_6}
\end{equation}
\item \textbf{Computing visual attributes}  Final attribute representation $i_{a} \in  \mathbb{R}^{58} (f_A(x))$ is computed by the equation (\ref{eqn:eqnB_7}) below, where the $e \in  \mathbb{R}^{d}$ indicates a feature vector of an input image.
\begin{equation}
i_{a} = Q^{t} \cdot e 
\label{eqn:eqnB_7}
\end{equation}
\end{enumerate}
\section{EVALUATION METRICS}
\begin{figure*}[h]
\centering
\begin{subfigure}[b]{0.495\textwidth}
\includegraphics[width=\textwidth]{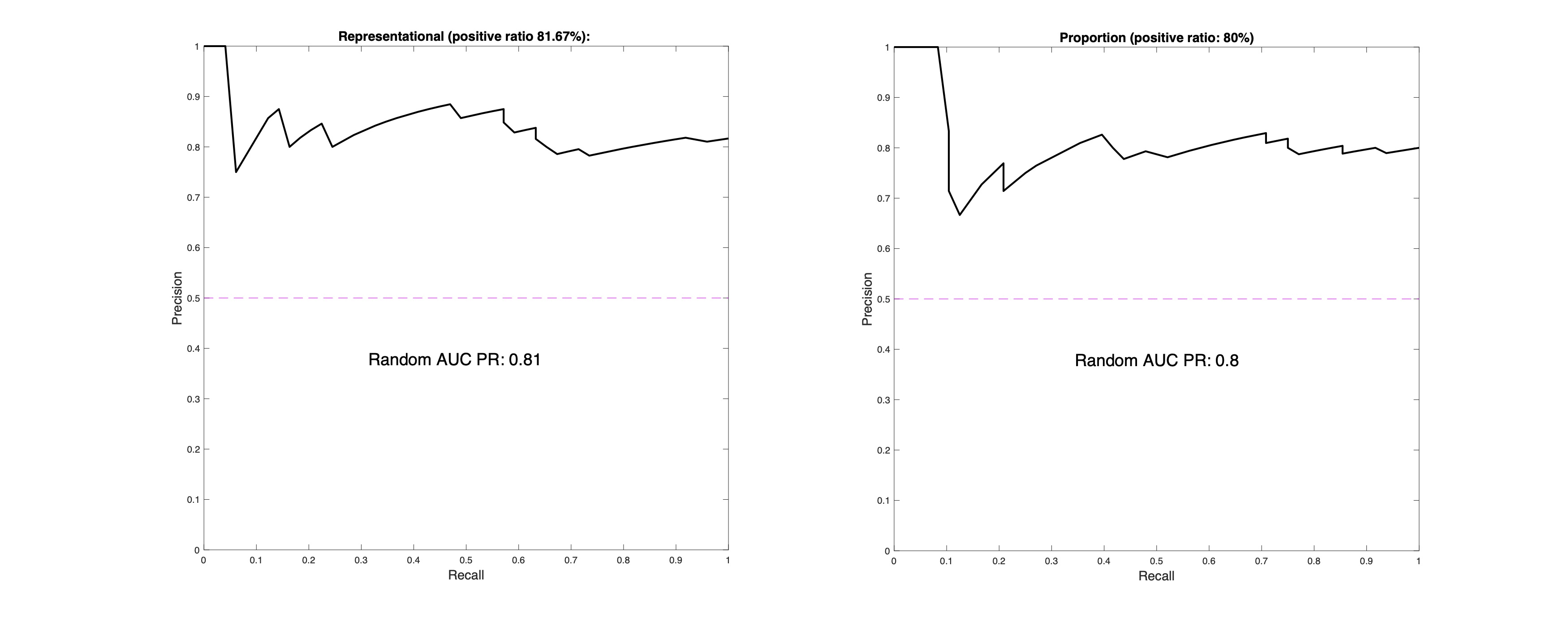}
\caption{PR  Example: ``representational'' and ``proportion''.}
\end{subfigure}
\begin{subfigure}[b]{0.495\textwidth}
\includegraphics[width=\textwidth]{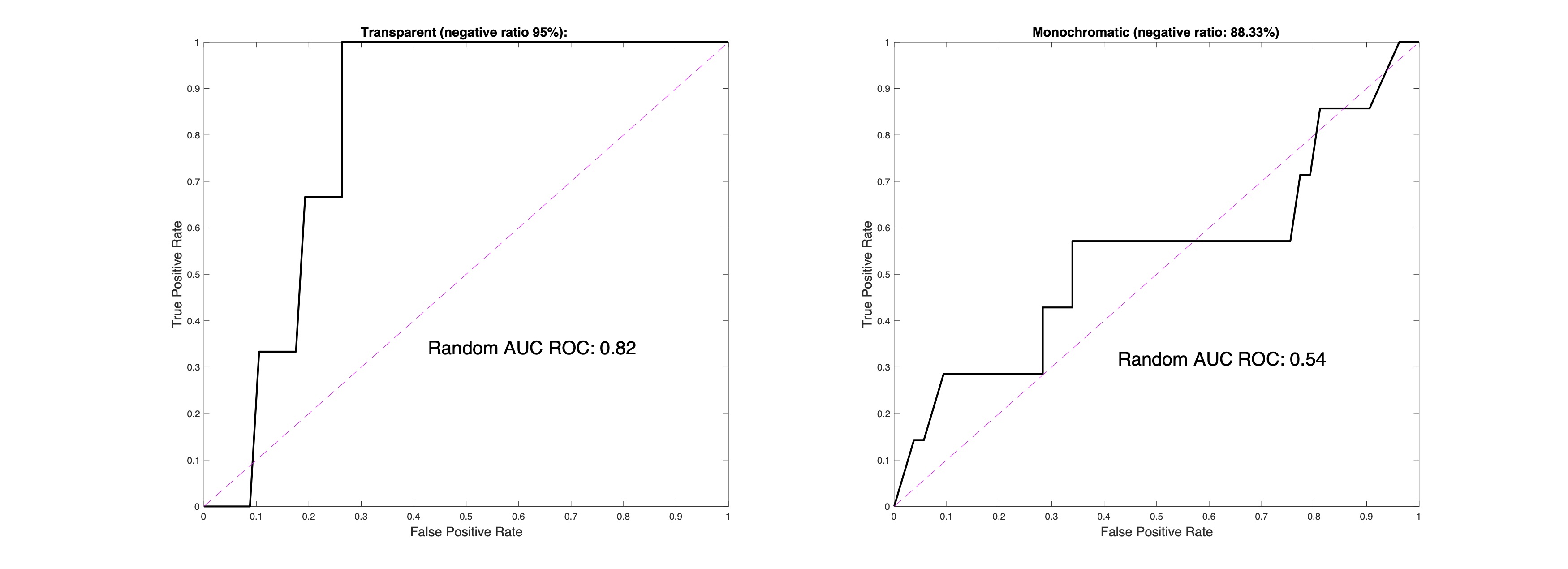}
\caption{AUC example:``transparent'' and ``monochromatic''}
\end{subfigure}
\caption{ Example cases of PR and AUC. AUC can be biased by the imbalance of large negative ground truths as the case of “transparent” in figure (b). However, we confirmed AUC is less affected by the imbalance of data in general than PR or mAP, while  Precision-Recall (PR)  presented deceptively too high score as shown in (a). For AUC, most of random cases scored around 0.5 regardless of the negative ratios like  ``monochromatic'' in figure (b).}
\label{fig:figC_1}
\end{figure*}
\begin{table*}[h!]
\centering
\resizebox{0.8\columnwidth}{!}{
\begin{tabular}{cc} 
\toprule
    \multicolumn{2}{c}{58 Visual Elements of Art}\\
    \cmidrule(r){1-2}
 Elements of Art &  Words \\
\midrule
Subject&representational (0.18/0.56), non-representational (0.76/0.5)\\\hline
Line&\makecell{blurred (0.72/0.43),
broken (0.82/0.43),
controlled (0.35/0.51)\\
curved (0.6/0.52),
diagonal (0.82/0.53),
horizontal (0.61/0.55)\\
vertical (0.46/0.45),
meandering (0.93/0.37)\\
thick (0.78/0.51),
thin (0.8/0.45),
active (0.53/0.48)\\
energetic (0.56/0.46),
straight (0.76/0.52)} \\\hline
Texture&\makecell{bumpy (0.87/0.42),
flat (0.45/0.53),
smooth (0.63/0.58)\\
gestural (0.63/0.46),
rough (0.68/0.44)
}\\\hline
Color&\makecell{calm (0.63/0.46),
cool (0.63/0.49),
chromatic (0.36/0.54)\\
monochromatic (0.88/0.54),
muted (0.8/0.35)\\
warm (0.5/0.56),
transparent (0.95/0.82)}\\\hline
Shape&\makecell{ambiguous (0.72/0.42),
geometric (0.78/0.49),
amorphous (0.88/0.4)\\
biomorphic (0.82/0.4),
closed (0.3/0.51),
open (0.68/0.46),
distorted (0.75/0.45)\\
heavy (0.72/0.52),
linear (0.72/0.51),
organic (0.76/0.43),
abstract (0.69/0.48)\\
decorative (0.63/0.56),
kinetic (0.77/0.48),
light (0.81/0.55)}\\\hline
\makecell{Light and Space}&\makecell{bright (0.56/0.48),
dark (0.74/0.57), atmospheric (0.68/0.54) \\
planar (0.58/0.55),
perspective (0.5/0.54)}\\\hline
\makecell{General
Principles\\of Art} &\makecell{overlapping (0.59/0.56),
balance (0.51/0.41),
contrast (0.47/0.49)\\
harmony (0.5/0.56),
pattern (0.61/0.56)
repetition (0.57/0.54)\\
rhythm (0.65/0.48),
unity (0.59/0.45),
variety (0.55/0.46)\\
symmetry (0.78/0.48),
proportion (0.35/0.51),
parallel (0.77/0.56)}\\         
\bottomrule
\end{tabular}}
\caption{58 Visual Elements: ratios of negative ground truth (value on the left-hand side) and random-AUC scores (value on the right-hand side) on the $120$ paintings in ``eval''. We confirmed AUC is less affected by the imbalance of data in general than PR or mAP. Most of random cases scored close to 0.5 regardless of the negative  ratios of ground truths.}
\label{tab:tabC_1}
\end{table*}
Both mean Average Precision (mAP) and area of Precision of Recall curve (PR)  were initially considered along with AUC. However, we found that some visual concepts were highly imbalanced in ``eval’' (positive samples $\gg$ negative samples), so the two metrics presented deceptively too high  scores. For example, for the concepts of ``representational'' and ``proportion'', randomly ordered samples achieved 0.81  and 0.8 PR  because more than 80\% paintings of the 120 paintings in ``eval’' are relevant or somewhat relevant to the concepts. In  Figure \ref{fig:figC_1} (a), each element's positive ground truth ratio is directly reflected on their PR scores. 
\par Theoretically,  AUC scores of random samples  can also be biased by larger negative ground truths as presented in the ``transparent'' in Figure \ref{fig:figC_1} (b). However, we confirmed  AUC is less affected by the imbalance  of data in general than PR or mAP (PR is the lower bound of mAP). As  we  compared  random  performances of AUC, in  most of cases the random scores are marked around 0.5 regardless of the negative ratios like  the ``monochromatic'' of Figure \ref{fig:figC_1} (b).  As a reference,  both ratios of negative ground truths and random-AUC scores $(\cdot/ \cdot)$ on ``eval'' are presented for all 58 visual elements of art in Table \ref{tab:tabC_1}.
\section{SUN and CUB AP Plots}
\begin{figure}
  \centering
  \includegraphics[width=0.95\textwidth]{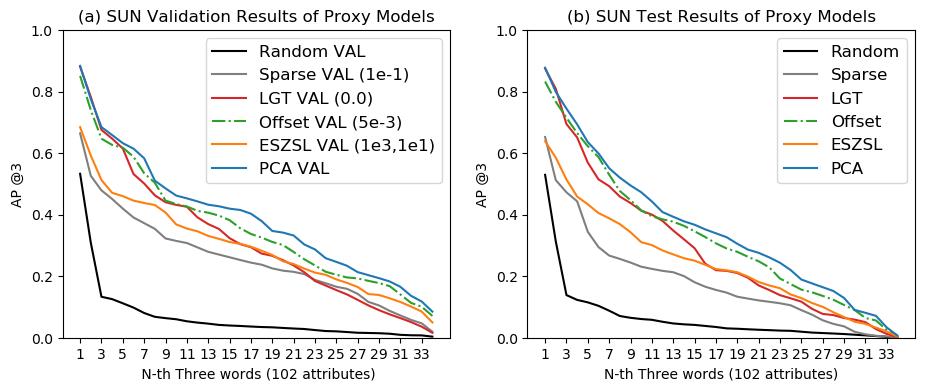}
  \caption{SUN AP performance of proxy models: Due to the sparsity of $G$ matrix of SUN, LGT  is degraded. The ranking relations among the five models in validation are also hold in test.}
  \label{fig:figD_1}
\end{figure}

\begin{figure}
  \centering
  \includegraphics[width=1.0\textwidth]{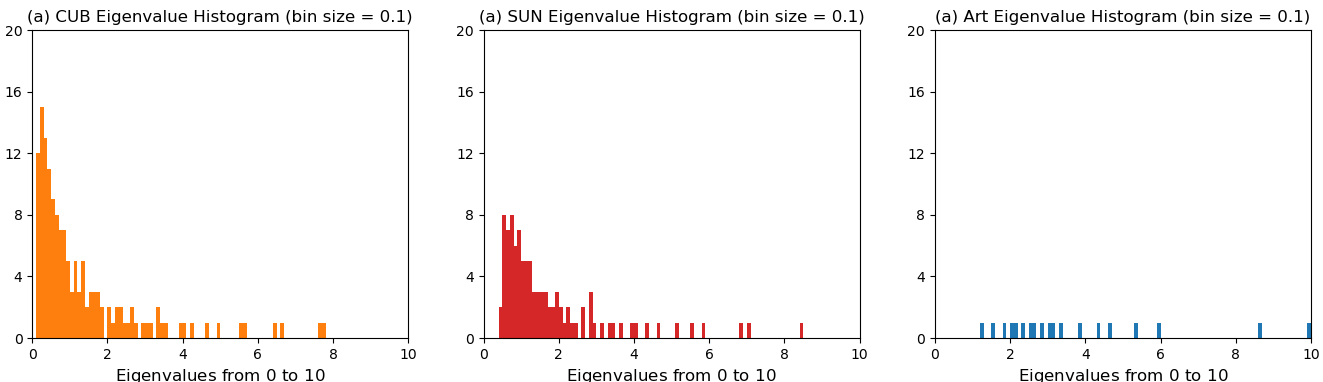}
  \caption{Eigenvalue-Distributions of CUB, SUN, and Art: The cropped eigenvalue-distributions   between 0.0 to 10.0 are presented for CUB, SUN, and Wikiart to show CUB's $G$ matrix has more small eigenvalues than others. This implies there exist subtle visual differences among  bird-classes.}
  \label{fig:figD_2}
\end{figure}
\begin{figure}
 \centering
  \includegraphics[width=0.95\textwidth]{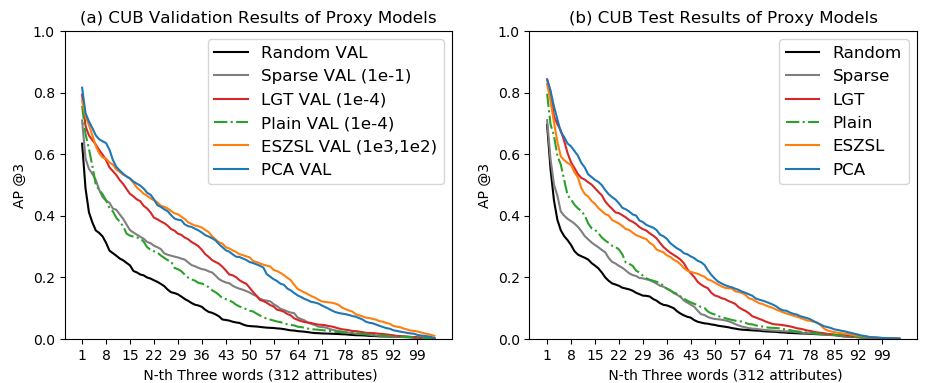}
  \caption{CUB AP performance of proxy models: Since the visual difference among bird-classes   are very subtle, fine, and local, Offset and SVD were not learnable.}
  \label{fig:figD_3}
\end{figure}
To understand the models in more general situations, SUN \cite{patterson2014sun} and CUB \cite{WahCUB_200_2011} are tested for all five proxy models. All experiments are based on the standard splits (train, validation, and test), proposed by Xian et al. \cite{xian2018zero} for zero-shot learning. The classes in the test split are unseen in training and the validation split is used to select hyper parameters. Since both ground truth attributes are imbalanced by very large negative samples (the mean  of all the samples is $0.065$ for SUN and $0.1$ for CUB at the binary threshold of $0.5$), mean Average Precision (AP) is used for evaluation. For $G$ matrix, their ground truth samples are averaged.  PCA scored the best on  both SUN and CUB but other models performed differently depending on the data sets. All AP plots for SUN and CUB are presented in Figure \ref{fig:figD_1} and Figure \ref{fig:figD_3}.  Eigenvalue-distributions of CUB, SUN, and Art are presented in Figure \ref{fig:figD_2}.
\end{document}